\documentclass{article}

\usepackage[usenames,dvipsnames]{xcolor}         
\usepackage[export]{adjustbox}  

\usepackage[preprint]{neurips_2024}

\usepackage[utf8]{inputenc} 
\usepackage[T1]{fontenc}    
\usepackage[colorlinks=True,citecolor=ForestGreen]{hyperref}       
\usepackage{url}            
\usepackage{booktabs}       
\usepackage{amsfonts}       
\usepackage{nicefrac}       
\usepackage{microtype}      

\usepackage{amsmath}
\usepackage{amssymb}
\usepackage{verbatim}

\usepackage[inline]{enumitem}
\usepackage[textsize=tiny,textwidth=2cm]{todonotes}
\usepackage{subcaption}

\newcommand{\ourtitle}{Cascade-Aware Training of Language Models}
\title{\ourtitle}

\author{%
  Congchao Wang \\
  Google Inc.\\
  \texttt{congchaowang@google.com} \\
  \And
  Sean Augenstein \\
  Google Inc.\\
  \texttt{saugenst@google.com} \\
  \And  
  Keith Rush \\
  Google Inc.\\
  \texttt{krush@google.com} \\
  \And  
  Wittawat Jitkrittum \\
  Google Inc.\\
  \texttt{wittawat@google.com} \\
  \And  
  Harikrishna Narasimhan \\
  Google Inc.\\
  \texttt{hnarasimhan@google.com} \\
  \And
  Ankit Singh Rawat \\
  Google Inc.\\
  \texttt{ankitsrawat@google.com} \\
  \And  
  Aditya Krishna Menon \\
  Google Inc.\\
  \texttt{adityakmenon@google.com} \\
  \And
  Alec Go \\
  Google Inc.\\
  \texttt{ago@google.com} \\
}

\begin{document}

\maketitle


\begin{abstract}
Reducing serving cost and latency is a fundamental concern for the deployment of language models (LMs) in business applications. To address this, \emph{cascades} of LMs offer an effective solution that conditionally employ smaller models for simpler queries. Cascaded systems are typically built with independently trained models, neglecting the advantages of considering inference-time interactions of the cascaded LMs during training. In this paper, we present \emph{cascade-aware training} (CAT), an approach to optimizing the overall quality-cost performance tradeoff of a cascade of LMs. We achieve inference-time benefits by training the small LM with awareness of its place in a cascade and downstream capabilities. We demonstrate the value of the proposed method with over 60 LM tasks of the SuperGLUE, WMT22, and FLAN2021 datasets.
\end{abstract}

\section{Introduction}
\label{sec.intro}

\begin{figure}
    \centering
    \includegraphics[width=\textwidth]{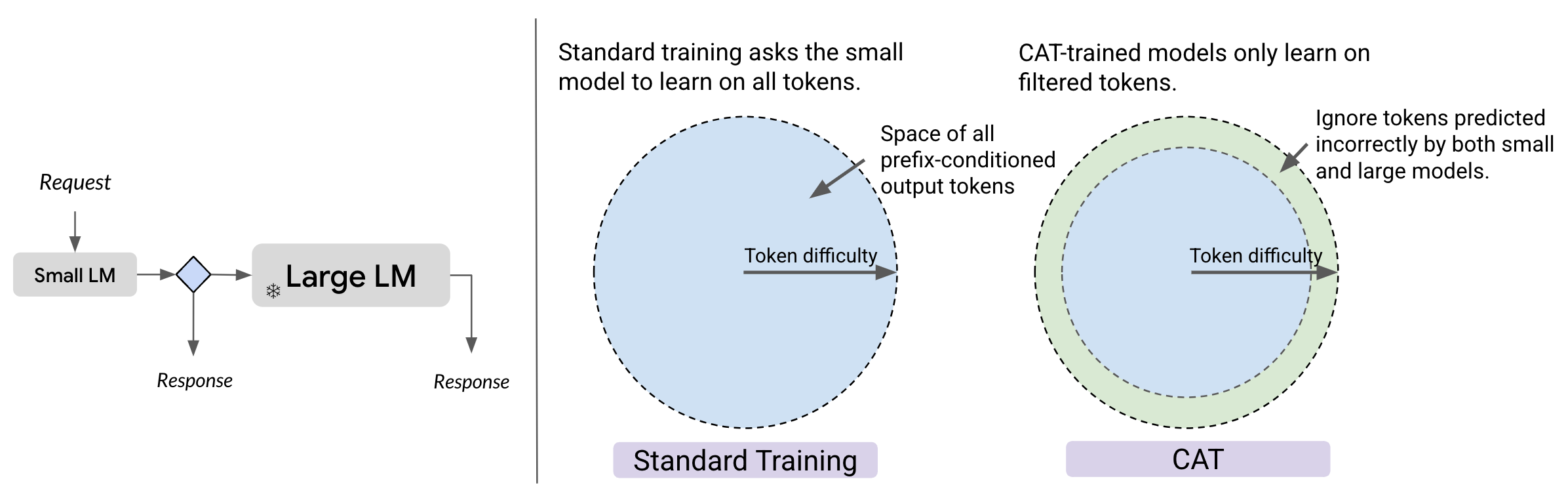}
    \caption{\textbf{Left:} Cascade setup at inference time. The small model is deployed along side the large model that guided its training.
    \textbf{Right:} Our proposed cascade-aware training (CAT). The small model has access to a trained, fixed, large model during training. 
    The proposed training objective (see \eqref{eq:cat_loss}) is a generalization of the standard one-hot cross entropy and KL-divergence based distillation loss, where losses are only accounted for on 
    tokens that are predicted correctly by the small or the large model (i.e., tokens that are not too difficult).
    }
    \label{fig:setup_and_cat}
\end{figure}

Dense neural networks, such as large language models (LLMs), incur significant computational cost to train and serve. Conditional computation---where a strict subset of model parameters are activated on some queries---is a manner of reducing cost. One general approach is to arrange a group of models into a \emph{cascade} of varying scales \citep{Viola:2001, Varshney:2022}, with the smallest being least capable (but computationally cheapest) and largest being most capable (but most expensive). A query is routed through the cascade, and uses the smallest model which is `confident' (in some concrete sense) to compute the response. See Figure~\ref{fig:setup_and_cat}, Left.

Such a modular arrangement naturally admits physically distributed deployments: e.g., a low-latency small model on a mobile device where queries originate, augmented by a high-latency larger model on a server in a datacenter \citep{rawat2021doubt, kag2023efficient}. Compared to sending all queries to the highest-quality model at the server, the cascade saves on both computation cost (e.g., average floating point operations or `FLOPs' per query, a proxy for power usage) and average query latency.

Cascades have been extensively studied for image classification and segmentation~\citep{Viola:2001,Trapeznikov:2013,Bolukbasi:2017,Huang:2018,Wang:2018,Streeter:2018,rawat2021doubt,wang2022wisdom,kag2023efficient,jitkrittum2023confbaseddef}, for classification-based natural language processing (NLP)~\citep{rawat2021doubt,Mamou:2022,Varshney:2022,Khalili:2022,Dohan:2022}, and recently for \emph{generative} NLP~\citep{Chen:2023a,gupta2024lmcascades,Yue:2024}. The latter's central challenge is that it involves variable-length \emph{sequence} outputs, where a suitable confidence measure that can indicate the quality of the response is non-obvious~\citep{gupta2024lmcascades}.

Typically, neural network cascades leverage preexisting models, dropping them into place in the cascade as is \citep{Varshney:2022}. Any considerations of the overall cascade's performance are \emph{post-hoc} and concentrated on aspects apart from the constituent models, like the routing or deferral logic (e.g., as in \citet{ narasimhan2022posthocest,jitkrittum2023confbaseddef,gupta2024lmcascades,Chen:2023a,Yue:2024}). The models themselves remain fixed and \emph{cascade-oblivious}. But intuitively it would be preferable if the small model did not expend capacity attempting to handle `hard' queries that will be routed onwards anyway. A few \emph{cascade-aware} approaches have been proposed in the literature for non-generative tasks. \cite{kag2023efficient} trains an entire cascaded system for image classification end-to-end. \cite{rawat2021doubt} uses carefully tuned distillation-inspired losses to isolate knowledge in a `lite' (i.e., small) model while keeping their large model frozen. Both show significant improvements, but are not immediately applicable to the generative LLM setting. 

To the best of our knowledge, the present work is the first to consider how to perform \emph{cascade-aware} training for cascades of \emph{LLMs}. Fine-tuning a constituent LLM in a cascade of LLMs requires addressing several underlying technical challenges. First, LLMs in their generative capacity are deployed in an autoregressive manner; inference routing decisions are made at the sequence level (as noted by~\citet{gupta2024lmcascades}). Training, however, takes place at the \emph{token} level. Bridging this gap between token-based training and sequence-level inference routing is not straightforward. Second, and related, the notion of `easy' or `hard' training content is complex as it must be judged at the token level. Finally, the larger model(s) may have several billions of parameters (with correspondingly massive training costs).

In this paper, we describe how to significantly improve a cascade of language models (LMs) by fine-tuning the smallest LM with `awareness' of its place in the cascade and the capabilities of the larger LMs that support it, in a scalable manner, while bridging the token-to-sequence training to inference gap. Such \emph{cascade-aware training} (as we refer to it) yields a significantly improved quality-cost tradeoff relation compared to various baselines, at relatively small increased training cost. Crucially, we only query the typical functional API of the larger model with respect to which we train, analogous to that exposed by ChatGPT~\citep{chatgpt} or Gemini~\citep{geminiteam2024gemini}. We seek a `sweet spot' where we impact the deferral behavior enough to significantly reduce inference-time computational cost, without dramatically increasing the computational cost of training, and while leaving the capabilities of the large model unmodified.

The contributions of this paper are as follows:
\begin{enumerate}[leftmargin=16pt,label=(\roman*)]
    \item We present a practical, scalable, and general method to improve the quality-to-cost tradeoff curve for a cascade of two language models (Section \ref{sec.cat_algos}). This method functions by leveraging the predictions of the large model to define a new loss function for the small model. Intuitively, the small model is encouraged to focus its capacity on `easier' examples, by selectively masking certain tokens--those on which both models' prediction was incorrect (see Figure~\ref{fig:setup_and_cat}, Right). 
    
    \item We observe that this altered loss function improves the cascade quality primarily by increasing performance of the small model when asked to process a significant fraction of the queries, often improving the performance of the small model when it processes \emph{all} queries (Section \ref{sec.exp}). In this sense, our methods can be understood as a a novel approach to token-wise data filtering without sacrificing large amounts of data.
    \item To the best of our knowledge, our work is the first to consider \emph{directly optimizing} language models for their position in cascades in a multi-task scenario encompassing both generation and classification tasks, and the first to demonstrate that \emph{token-level} loss adjustments can be translated to \emph{sequence-level} cascade gains.
\end{enumerate}
We demonstrate all of the above with experiments on the SuperGLUE \citep{wang2019superglue}, WMT22 \citep{kocmi2022wmt}, and FLAN2021 \citep{wei2021flan} datasets encompassing over 60 LM tasks.

\section{Setup}
\label{sec.setup}

\paragraph{Language Models (LMs)}

Let $\mathbf{x}:=(x_{1},\ldots,x_{M})\in\mathcal{X}$ denote an input sequence (or `query') where each token $x_{i}$ is an element from a vocabulary $\mathcal{V}:=[V]:=\{1,\ldots,V\}$ of cardinality
$V$, and $\mathcal{X}:=\mathcal{V}^*$. Similarly, let $\mathbf{y}:=(y_{1},\ldots,y_{N})\in\mathcal{Y}:=\mathcal{V}^*$
be an output sequence (or `response') where each output token $y_{i}\in\mathcal{V}.$
A language model $p$ is a parametric probability distribution that
computes the probability $p(\mathbf{y}|\mathbf{x})$ of an output
sequence $\mathbf{y}$ given an input sequence $\mathbf{x}$. This
conditional probability may be expressed as:
\begin{align}
p(\mathbf{y}|\mathbf{x}) & =\prod_{i=1}^{N}p(y_{i}|\mathbf{x},\mathbf{y}_{<i}),
\label{eq:autoregressive}
\end{align}
where $\mathbf{y}_{<i}:=(y_{1},\ldots,y_{i-1})$ and $y_{<1}=\emptyset$.
Expressing the probability in this auto-regressive manner allows one to parameterize a model to repeatedly predict one output token at a time. A common modeling approach is to set $p(\cdot|\mathbf{x},\mathbf{y}_{<i})=\mathrm{softmax}\left(f(\mathbf{x},\mathbf{y}_{<i})\right)$ where $f(\mathbf{x},\mathbf{y}_{<i})\in\mathbb{R}^{V}$ is a vector of per-token logit scores. Common network architecture choices for modeling $f$ are Recurrent Neural Networks, such as Long Short-Term Memory \citep{hochreiter1997lstm}, or 
Transformers \citep{vaswani2017transformer}. The latter are generally considered the state-of-the-art, and we will use them in our experiments in Section~\ref{sec.exp}.

\paragraph{LM Training Losses}
\label{sec:train_next_tokens}Suppose that we observe a training set
$\{(\mathbf{x}^{(j)},\mathbf{y}^{(j)})\}_{j=1}^{S}$ containing $S$ query-response pairs. With the auto-regressive representation in
\eqref{eq:autoregressive}, a common way to use this training set to fine-tune the LM is by minimizing the average over examples of (one-hot) cross-entropy loss, which is the sum negative log likelihoods of the $N$ output tokens of a particular example $(\mathbf{x}, \mathbf{y})$:

\begin{align}
L_{\mathrm{xent}}(\mathbf{x},\mathbf{y}) & = - \sum_{i=1}^{N}\log p(y_{i}|\mathbf{x},\mathbf{y}_{<i}).
\label{eq:onehot_ce}
\end{align}

This can be augmented with \emph{distillation} \citep{Bucilua:2006,hinton2015distillation} in some proportion $w \in [0, 1]$, if we have a well-trained (teacher) LM, $p_{\mathrm{teach}}$, available to `teach' the (student) LM undergoing training, $p$:

\begin{align}
L_{\mathrm{dist}}(\mathbf{x},\mathbf{y}) & =-\sum_{i=1}^{N}\Bigl( w \cdot \log p(y_{i}|\mathbf{x},\mathbf{y}_{<i}) + (1-w) \cdot \sum_{y'=1}^{V}p_{\mathrm{teach}}(y'|\mathbf{x},\mathbf{y}_{<i})\log p(y'|\mathbf{x},\mathbf{y}_{<i}) \Bigr)
 \label{eq:distill_ce}
\end{align}
The latter involves \emph{token-level} distillation~\citep{sanh2020distilbert}, as opposed to \emph{sequence-level} distillation~\citep{kim-rush-2016-sequence,gu2024minillm,wei2024sentencelevel}. Recent work has also explored the utility of modifying the samples on which token-level distillation is performed~\citep{agarwal2024onpolicy}.

The above standard approaches (\eqref{eq:onehot_ce} and \eqref{eq:distill_ce}) consider all output tokens when calculating the loss. This penalizes the LM equally for its prediction ability on \emph{all} tokens. In Section~\ref{sec.cat_algos}, we will present an alternative approach which \emph{selectively} considers which tokens should count when calculating loss.

\paragraph{LM Cascades}

Let $p_{S}$ and $p_{L}$ be a small model and a large model, respectively. At test time, given an input query $\mathbf{x}$, we call a pre-determined deferral rule $r\colon\mathcal{X}\to\mathbb{R}$ to obtain a routing score (or deferral score)
$r(\mathbf{x})$. The small model $p_{S}$ is used if $r(\mathbf{x})<\tau$
and the large model is used otherwise, where $\tau\in\mathbb{R}$
is a threshold to be specified. The cascade model may be written succinctly as:
\begin{align}
p_{\rm cas}(\mathbf{y}|\mathbf{x}) & :=1[r(\mathbf{x})<\tau] \cdot p_{S}(\mathbf{y}|\mathbf{x}) + 1[r(\mathbf{x})\ge\tau] \cdot p_{L}(\mathbf{y}|\mathbf{x}).
\label{eq:cascade}
\end{align}

There are many design choices for the function $r$ (see \cite{gupta2024lmcascades,wang2022wisdom}). 

In this work, we focus on a deferral arrangement where $r$ depends on $p_S$. That is, each input query $\mathbf{x}$  always triggers a call to $p_S$, and then $r$ is used to decide whether to additionally invoke (or `defer' to) $p_L$. In particular, we use the confidence (as measured by normalized log-likelihood of the output sequence) of $p_S$ to decide whether to use $p_S$ for prediction or use $p_L$.
Concretely, let $\mathbf{y}_{S}\sim p_{S}(\cdot|\mathbf{x})$
be the output sequence consisting of $N$ tokens. The deferral rule of interest is:
\begin{align}
r(\mathbf{x})=-\frac{1}{N}\sum_{i=1}^{N}\log p_{S}(y_{S,i}|\mathbf{x},\mathbf{y}_{S,<i}).
\label{eq:routing}
\end{align}

\paragraph{Cascade Cost}
The point of a model cascade is to reduce costs, like expected number of floating point operations (FLOPs) per query or expected latency per query, via conditional activation of parameters. If the cost to serve query $\mathbf{x}$ is $C_{S}(\mathbf{x})$ for the small model and $C_{L}(\mathbf{x})$ for the large model, the cost to serve query $\mathbf{x}$ by the overall neural cascade is:
\begin{align}
C_{\rm cas}(\mathbf{x}) = C_{S}(\mathbf{x}) + 1[r(\mathbf{x})\ge\tau] \cdot C_{L}(\mathbf{x})
\label{eq:cost}
\end{align}


\section{Cascade-Aware Training (CAT)}
\label{sec.cat_algos}

Given the preliminaries in the previous section, we can describe the exact problem we will optimize. We assume that $p_{L}$ has already been fine-tuned on desired tasks and is fixed. We are interested in fine-tuning $p_{S}$ in a manner that, when deployed jointly alongside $p_{L}$ in an overall cascade $p_{\rm cas}$, achieves the best expected quality of response from \eqref{eq:cascade} for expected cost expended from \eqref{eq:cost}. 

Adjusting the parameters of $p_{S}$ can affect the quality-cost tradeoff in two ways: via changes to the small model's prediction accuracy, and via changes to the small model's prediction confidence (which affects routing decisions via \eqref{eq:routing}). Note that the routing threshold $\tau$ is a free parameter which controls the operating location on the quality-cost tradeoff curve. Cascade-aware training should provide a more desirable such tradeoff curve.

We now present our approach for training the small model $p_{S}$ in a cascade-aware manner. 
Given an input-output training example $(\mathbf{x}, \mathbf{y})$, we define the \emph{cascade-aware training loss} as:

\begin{align}\label{eq:cat_loss}
L_{\mathrm{cat-dist}}(\mathbf{x},\mathbf{y})  :=-\sum_{i=1}^{N}&\alpha_{i}\cdot \Bigl(w \cdot \log p_{S}(y_{i}|\mathbf{x},\mathbf{y}_{<i}) \\
&+ (1-w) \cdot \sum_{y'=1}^{V}p_{L}(y'|\mathbf{x},\mathbf{y}_{<i})\log p_S(y'|\mathbf{x},\mathbf{y}_{<i})\nonumber\Bigr),
\end{align}

\begin{align}\label{eq:alpha_def}
\alpha_i = \alpha_{i}(\mathbf{x},\mathbf{y}_{<i}) & :=1\left[y_i=\arg\max_{y' \in \mathcal{V}}p_{S}(y'|\mathbf{x},\mathbf{y}_{<i}) \lor y_i=\arg\max_{y' \in \mathcal{V}}p_{L}(y'|\mathbf{x},\mathbf{y}_{<i})\right].
\end{align}

This loss is nearly the same as $L_{\mathrm{dist}}$ in \eqref{eq:distill_ce}, with the large model leveraged as teacher ($p_{\mathrm{teach}}=p_{L}$). The difference is that we selectively ignore some tokens (those wrongly predicted by both the small and large models), via the $\alpha_i$ term (see Figure~\ref{fig:setup_and_cat} Right for illustration). This focuses the small model to learn to predict learnable output tokens, where a token is considered learnable if it can be predicted correctly by one of the two models. We can also consider a simplified version of cascade-aware training loss when $w=1$, i.e. a cascade-aware variant of just the one-hot cross-entropy loss: $L_{\mathrm{cat-xent}}$. Note that $\alpha_i$ is non-differentiable, so it influences model updating only by selectively ignoring certain tokens. 

The token filtering criterion defined by $\alpha_i$ serves to consider only tokens that are not too difficult, i.e., those predicted correctly by at least one model. This approach allows the small model to optimize its limited capacity by focusing on learnable tokens curated by the large model, which helps improve both the small model's overall accuracy and its output confidence measure. Intuitively, this strategy leverages the large model's capability to concentrate the small model on a learnable token set (Figure~\ref{fig:setup_and_cat}, Right), thereby enhancing the small model's confidence and accuracy. Simultaneously, it maintains exploration by including tokens that the small model can correctly predict. This overall confidence enhancement on the correctly predicted tokens increases the reliability of the score from \eqref{eq:routing}, serving as a robust deferral indicator. As shall be seen in Section \ref{sec:exp:improve-non-defer-example}, a CAT-trained model exhibits improved accuracy on examples where it is confident.



Since $\alpha_i$ acts token-wise, it can be easily applied to the training or fine-tuning of language models. \citet{rawat2021doubt} demonstrated that filtering out the `hard' sequence (the whole training example)
can also benefit the cascade, allowing the small model to learn more efficiently on `easy' ones. However, designing an appropriate criterion for `hard' sequences (compared to `hard' tokens) is non-trivial. Additionally, even within difficult sequences, some tokens may not be inherently difficult to predict. Discarding entire sequences could result in the loss of valuable training tokens. 

The experiments in the next section will show the benefits to the cascade of fine-tuning the small model in a cascade-aware manner.

\section{Experiments}
\label{sec.exp}
In this section, we present a series of experiments comparing the performance of small models trained with cascade-aware training (CAT) against models trained with the more commonly used cross-entropy loss and distillation loss. These experiments were conducted on three datasets encompassing over 60 language model tasks. Additionally, we provide a case study, demonstrating that small models trained with CAT loss exhibit improved accuracy on examples where they have high confidence.
\subsection{CAT improves performance of the cascade}

\paragraph{Datasets}
To verify our hypothesis that CAT could benefit the cascade with more robust deferral indicator from the small model, we conducted fine-tuning experiments on the SuperGLUE \citep{wang2019superglue}, WMT22 \citep{kocmi2022wmt}, and FLAN2021 \citep{wei2021flan} datasets, which encompass over 60+ language model tasks commonly used for multi-task language model fine-tuning. SuperGLUE and WMT22 datasets consist of classification and generation tasks, respectively. Given the wide range of languages in WMT22, we randomly selected three language pairs (en $\leftrightarrow$ zh, en $\leftrightarrow$ ja, and en $\leftrightarrow$ ru), resulting in six translation tasks. The performance measurements for SuperGLUE and WMT (accuracy and BLEU score) are well-accepted benchmarks for robust performance comparison. FLAN2021 datasets include a large number of tasks covering both classification and generation task types, allowing us to examine CAT's influence on both when trained together.

Beyond fine-tuning, we also explored the benefits of CAT on the pretraining task using the C4 datasets \citep{raffel2020c4t5}, as detailed in Appendix \ref{app.cat4pretrain}.

\paragraph{Models}
We employed PALM-2 language models \citep{anil2023palm}
as our cascading LM candidates. For all experiments, we used pretrained PALM-2 Gecko as the small LM and PALM-2 Otter as the large LM. For each model-dataset pair, we conducted a round of fine-tuning to ensure they performed reasonably well. {Note that for WMT22, the large model was fine-tuned on the entire WMT22 dataset, not just the three selected language pairs of interest. This make the gap between small and large models clearer.} Across all three datasets, the large model consistently outperformed the small model. We used greedy decoding for all LMs. Detailed training configurations, including learning rate, training steps, and batch size, are provided in Appendix \ref{app.hyperparam}.

\begin{figure}[t]
    \centering
    \begin{minipage}[b]{0.49\textwidth}
        \centering
        \includegraphics[width=\textwidth]{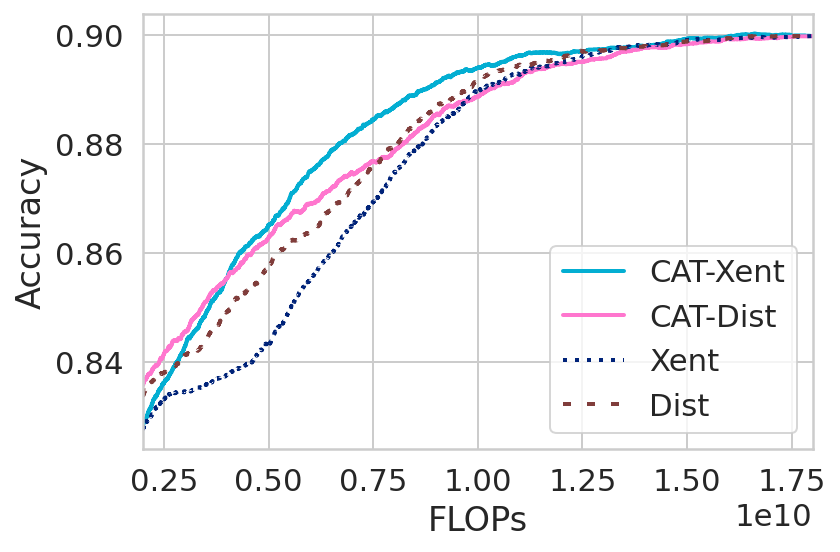}
        \subcaption{SuperGLUE}
    \end{minipage}
    \hfill
    \begin{minipage}[b]{0.49\textwidth}
        \centering
        \includegraphics[width=\textwidth]{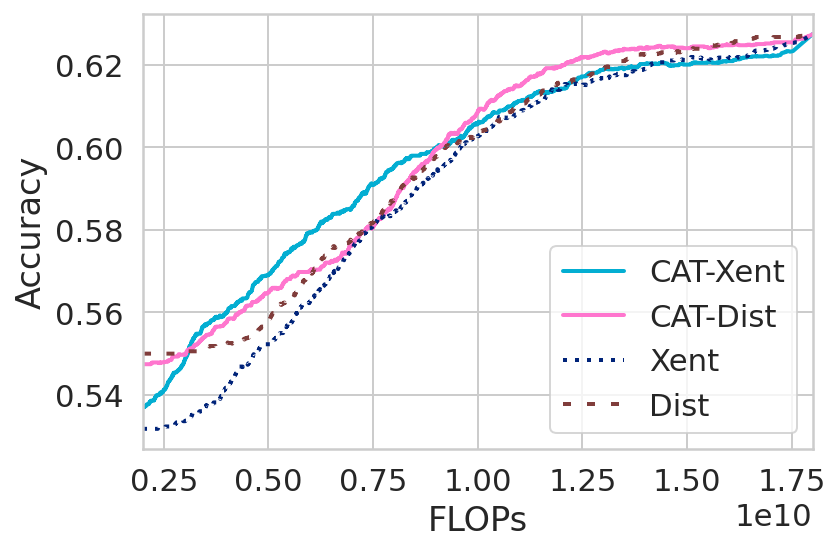}
        \subcaption{FLAN2021-Cls-Tasks}
    \end{minipage}
    \caption{Quality-vs.-FLOPs curves for classification tasks on SuperGLUE, and FLAN2021 datasets. We utilized two most commonly used training losses as comparison: cross-entropy ({\tt Xent}) and distillation with KL divergence ({\tt Dist}). Results showed that CAT benefits the cascade dramatically on both losses. On the SuperGLUE dataset as shown in (a), {\tt CAT-Xent} reduces 50\% FLOPs given fixed 86\% accuracy requests (13\% deferral request less). With the same 5 billion FLOPs budget, {\tt CAT-Xent} achieves 2\% absolute accuracy improvement. Meanwhile, CAT does not clearly downgrade the small model’s original capability (left end of the curves). When applying CAT with distillation, we can still see benefits for the cascade majority of the time ({\tt CAT-Dist} vs. {\tt Dist}). Similar benefits are also seen on FLAN2021 datasets in (b).}
    \label{fig:main_flops_acc_results}
\end{figure}
\begin{figure}[t]
    \centering
    \begin{minipage}[b]{0.49\textwidth}
        \centering
        \includegraphics[width=\textwidth]{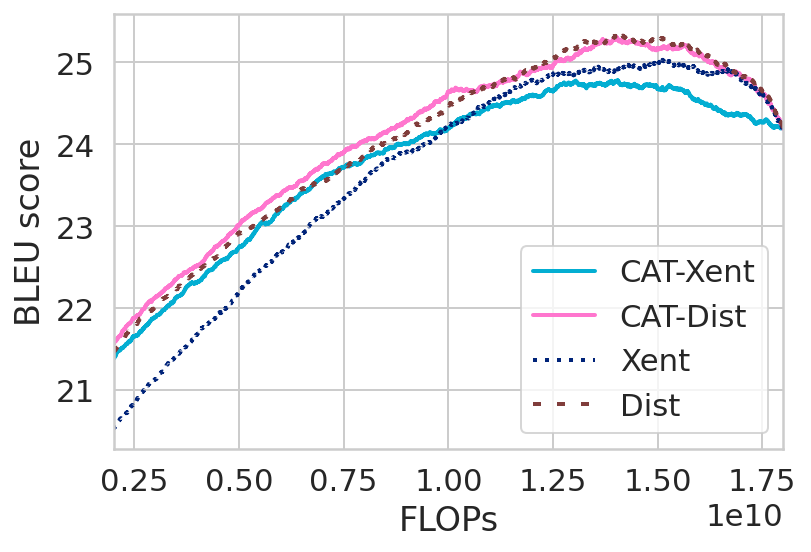}
        \subcaption{WMT22}
        \label{fig:res_wmt22}
    \end{minipage}
    \hfill
    \begin{minipage}[b]{0.49\textwidth}
        \centering
        \includegraphics[width=\textwidth]{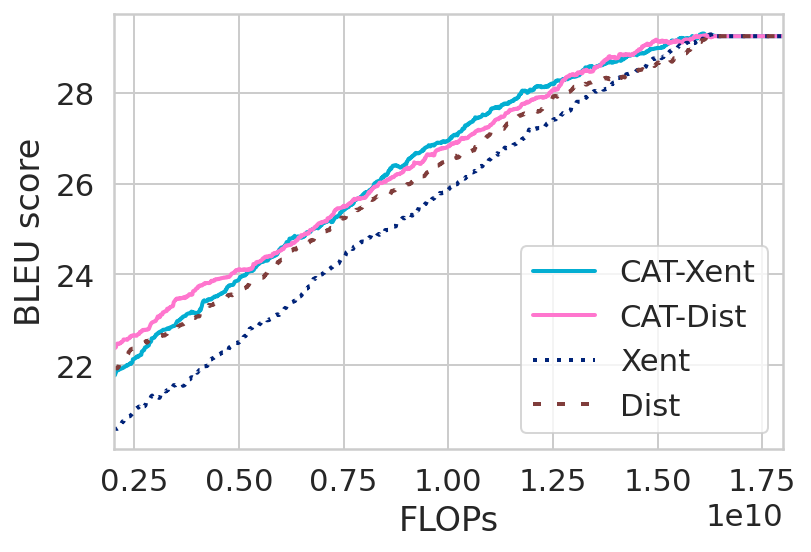}
        \subcaption{FLAN2021-Gen-Tasks}
        \label{fig:res_flan_gen}
    \end{minipage}
    \caption{Quality-vs-FLOPs curves for generation tasks on WMT22, and FLAN2021 datasets. We see benefits from CAT for the cascade in two aspects. One is the benefits to the small model’s quality directly and the other one is to the cascade. On both WMT22 and FLAN2021 datasets, we see clear higher BLEU scores from the small model trained with CAT and cross-entropy loss (see the starting point of {\tt CAT-Xent} vs. {\tt Xent} in (a) and (b)). This even mitigates the improvements from distillation ({\tt CAT-Xent} vs. {\tt Dist}). For cascade, we can also see the benefits, though not as significant as those on classification tasks.}
    \label{fig:main_flops_bleu_results}
\end{figure}

\paragraph{Evaluation}
We conducted evaluations on the benchmarks in a task-agnostic manner, wherein queries from different tasks were input into the language models. This approach is more realistic for language models with multi-task support. The results of task-specific comparisons are available in Appendix \ref{app.task_specific}. We used deferral curves (Accuracy-vs.-FLOPs or BLEU score-vs.-FLOPs) to summarize the models' cascading performance, following the methodology from previous works \citep{gupta2024lmcascades, jitkrittum2023confbaseddef, kag2023efficient}. FLOPs per token was used as the cost of interest (x-axis in Figures~ \ref{fig:main_flops_acc_results} and ~\ref{fig:main_flops_bleu_results}). We consider the per-token FLOPs cost of calling either constituent LM for inference as a constant (i.e., independent of query $\mathbf{x})$; we approximated it as twice the number of parameters in that LM (as in \citet{kaplan2020scaling} and \citet{hoffmann2022chinchilla}). Calculating the FLOPs cost in this way makes it essentially the same as the "deferral ratio" used in \citep{gupta2024lmcascades, jitkrittum2023confbaseddef, kag2023efficient}. We constructed the deferral curves by sweeping different thresholds $\tau$ in \eqref{eq:cost} on the deferral score $r(x)$ in \eqref{eq:routing}.

The FLAN2021 datasets provided 15 different metrics for evaluation. Here, we present results from 39 tasks whose evaluation metric is accuracy, and 8 tasks that rely on BLEU scores.

\paragraph{Baselines}
As CAT is applied on losses in a token-wise manner, we can easily apply it to various training paradigms for language models. We chose two loss functions for this study. One is the plain cross-entropy loss {\tt Xent} ($L_{\mathrm{xent}}$ in \eqref{eq:onehot_ce}), which is most commonly used in model fine-tuning. The other is model distillation loss {\tt Dist} ($L_{\mathrm{dist}}$ in \eqref{eq:distill_ce}). For the distillation experiments, we set $w=0.5$, which has proven reliable based on our experience. We applied cascade-aware training to these two losses and got our new design {\tt CAT-Xent} and {\tt CAT-Dist} (see $L_{\mathrm{cat-xent}}$ and $L_{\mathrm{cat-dist}}$ in \eqref{eq:cat_loss} in Section \ref{sec.cat_algos}). We also explored four other loss designs which are similar to \eqref{eq:cat_loss} but only based on \emph{either} $p_S$ or $p_L$; see our ablation study in Appendix \ref{apdx:sec.ablation_loss_design}. \citet{kag2023efficient,gupta2024lmcascades} are also compared with some adjustment to adapt the multi-task LM cascade, but their performance is not on-par with the other baselines. For the sake of clarity, we defer their presentation to Appendix \ref{app.other_baselines}.

\paragraph{Comparison: classification tasks}  Figure~\ref{fig:main_flops_acc_results} shows that CAT benefits the cascade dramatically on the SuperGLUE dataset, especially when the model is trained with one-hot cross entropy ({\tt CAT-Xent}). Compared with the training with plain one-hot cross entropy ({\tt Xent}), {\tt CAT-Xent} reduces 13\% FLOPs (large model calls) given fixed 87\% accuracy. 

With fixed 2 billion total FLOPs budget ($\sim$20\% large model calls), {\tt CAT-Xent} gets 2\% accuracy improvement. Meanwhile, CAT does not clearly downgrade the small model’s original capability (left end of the curves). When applying CAT with distillation, we can still see benefits for the cascade majority of the time ({\tt CAT-Dist} vs. {\tt Dist}). Similar trends can be seen on FLAN2021 datasets. It is worth mentioning that on both SuperGLUE and FLAN2021 dataset, distillation improves the small LM’s quality ({\tt Dist} vs. {\tt Xent}) compared to with one-hot cross entropy, {\tt CAT-Xent} levels up the cascade performance and out-performs the cascade with small LM fine-tuned with distillation, especially in the low-FLOPS range.

\paragraph{Comparison: generation tasks} Generation tasks commonly contain dozens to hundreds of tokens in the responses. Though more challenging than classification tasks, we see from Figure~\ref{fig:main_flops_bleu_results}
benefits from CAT for the cascade. 
One obvious gain from CAT is that the small LMs could enjoy better intrinsic quality with cross-entropy loss (higher BLEU scores see the starting points of {\tt CAT-Xent} versus {\tt Xent} in Figure~\ref{fig:main_flops_bleu_results}(a) and (b)) on WMT22 dataset. This improvement even mitigates the benefits gained from distillation ({\tt CAT-Xent} versus {\tt Dist}). This is further confirmed on the FLAN2021 dataset. We hypothesize that this improvement partially comes from the lower ratio of noisy training samples after token-wise filtering from CAT. In addtion, we can also see the benefits from CAT for the cascade on the generation tasks (e.g. {\tt CAT-Xent} versus {\tt Xent} on FLAN2021 or WMT22 in the low-to-intermediate deferral regime). 






\subsection{CAT improves accuracy on non-deferred examples}
\label{sec:exp:improve-non-defer-example}
In this section, we show that the model trained with the proposed CAT objective of \eqref{eq:cat_loss} has improved accuracy on examples on which it is confident (compared to a standard trained model).

Let $M$ be a task-specific metric function such that $M(\mathbf{x}, \mathbf{y}, p)$ gives a quality score of a predicted output sequence produced by a model $p$ relative to the ground-truth output $\mathbf{y}$. 
For instance, for SuperGLUE, a common choice is to set $M$ to be the 0-1 correctness indicator function.
The mean quality score $M_{\rm cas}(\tau)$ of a cascade of $p_S$  and $p_L$ is given by
%
\begin{align}
M_{\rm cas}(\tau) & =\mathbb{E}_{(\mathbf{x},\mathbf{y})}\left[M(\mathbf{x}, \mathbf{y},p_S)1[r(\mathbf{x}) < \tau]\right]
+\mathbb{E}_{(\mathbf{x},\mathbf{y})} \left[M(\mathbf{x}, \mathbf{y},p_L)1[r(\mathbf{x}) \ge \tau]\right] \label{eq:acc_decompose} \\ 
 & :=A_{1}(\tau) +A_{2}(\tau), \nonumber
\end{align}
where $\mathbb{E}$ may be replaced with an empirical expectation. 
We recall from \eqref{eq:routing} that $r(\mathbf{x})=-\frac{1}{N}\sum_{i=1}^{N}\log p_{S}(y_{S,i}|\mathbf{x},\mathbf{y}_{S,<i})$, where  $\mathbf{y}_{S}\sim p_{S}(\cdot|\mathbf{x})$ is the output generated from $p_S$.
In \eqref{eq:acc_decompose}, $A_1(\tau)$ thus denotes the average (unnormalized) quality score on examples that invoke $p_S$ as a function of the deferral threshold $\tau$. 
Similarly, $A_2(\tau)$ denotes the average quality score on examples that are deferred to $p_L$.
Note that the $y$-axis in Figure~\ref{fig:main_flops_bleu_results} shows $A_1(\tau)+A_2(\tau)$ as $\tau$ varies. 

\begin{figure}[t]
    \centering
    \begin{minipage}[b]{0.45\textwidth}
        \centering
        \includegraphics[width=0.99\textwidth]{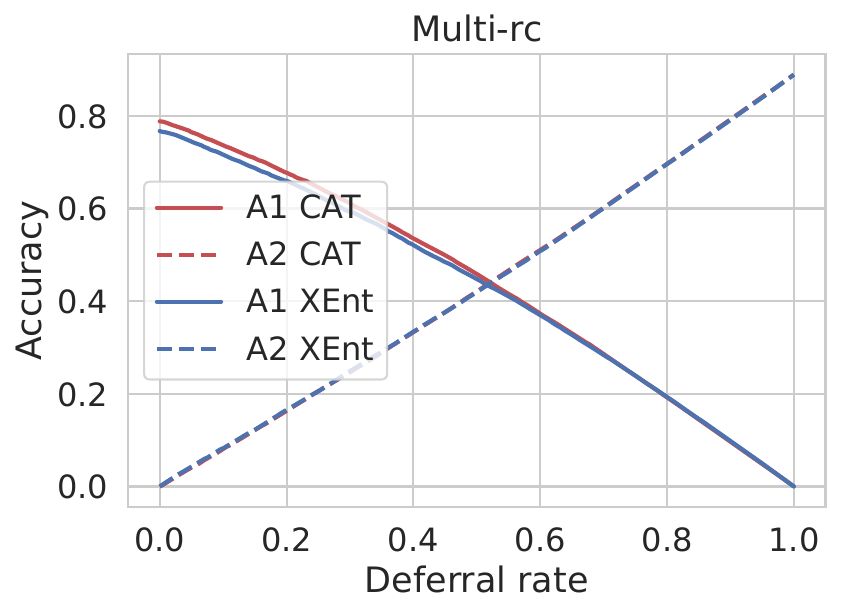}
        \subcaption{SuperGLUE Multi-RC}
    \end{minipage}
    \begin{minipage}[b]{0.45\textwidth}
        \centering
        \includegraphics[width=0.99\textwidth]{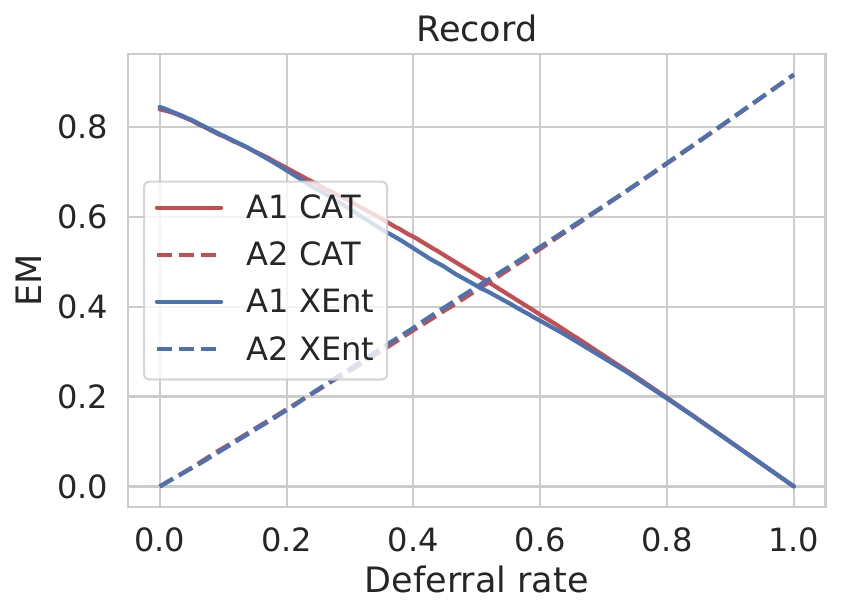}
        \subcaption{SuperGLUE Record}
    \end{minipage}
    \caption{Average quality scores on examples deferred to the small model (A1) and to the large model (A2). Accuracy numbers weighted by deferral rate, so that the sum of the two curves represents the accuracy of the cascaded system. We compare two models fine-tuned on SuperGLUE. \texttt{CAT} refers to fine-tuning using cascade-aware training loss of  \eqref{eq:cat_loss} (setting $w=1$), and \texttt{XEnt} refers to using the standard cross-entropy loss of \eqref{eq:onehot_ce}. We observe that the gains in accuracy is from A1. Note that curve of \texttt{A2-XEnt} is fully covered by \texttt{A2-CAT}
    }
    \label{fig:a1a2_analysis}
\end{figure}

Figure~\ref{fig:a1a2_analysis} shows $A_1(\tau)$ and $A_2(\tau)$ as we vary $\tau$, which in turn varies the deferral rate (i.e., the fraction of examples sent to $p_L$).
We compare two models fine-tuned on SuperGLUE: (i) CAT, and (ii) the model trained with the standard one-hot cross entropy. 
We observe that both models have roughly the same $A_2$ across all deferral rates, while the CAT-trained model has higher $A_1$. This indicates that on the set of examples each candidate model is confident, the CAT-trained model has higher accuracy.

We note that Figure~\ref{fig:a1a2_analysis} demonstrates two separate phenomena leading to improved performance of CAT-trained models. On SuperGLUE Multi-RC, 
$A_1$-\texttt{CAT} starts above $A_1$-\texttt{Xent} 
when deferral rate is 0; this indicates that $A_1$-\texttt{CAT} is simply a better model than $A_1$-\texttt{Xent}--all the queries are routed to the small model. Therefore $A_1$-\texttt{CAT} must have learned to correctly process some examples which $A_1$-\texttt{Xent} cannot. In the case of SuperGLUE Record, on the other hand, the two models start effectively overlapping on the left-hand side of the figure; it is only in the body of the deferral rate graph that $A_1$-\texttt{CAT} achieves separation from $A_1$-\texttt{Xent}.


    %

\section{Related work}
\label{sec.rw}

Having reviewed cascades and language models in Sections~\ref{sec.intro} and \ref{sec.setup}, and our novel contributions to cascade-aware training of language models in Sections~\ref{sec.cat_algos} and \ref{sec.exp}, we now describe a few other, more tangentially-related research directions.

\paragraph{Conditional Compute}

The deep learning revolution has been dominated by those models which most effectively leverage compute resources, a phenomenon famously characterized as the `bitter lesson'~\citep{sutton2019bitter}. One natural direction to better leverage compute resources (or, alternatively, expand model capacity for fixed compute cost) is to introduce conditional computation into the model architecture. Model cascades are one manner of doing this, making the inference computation conditional/adaptive.

A closely related technique for adaptive inference is \emph{early-exiting}, wherein a \emph{single} model is partitioned into multiple sub-models (typically via attaching classification heads to intermediate layers)~\citep{Teerapittayanon:2016,Huang:2018,Schwartz:2020,Xin:2020,jazbec2023towards}. As with a cascade, these less compute-intensive, intermediate sub-models can be invoked for `easy' queries. Recent works have successfully extended this paradigm to generative language models~\citep{schuster2022calm,kusupati2022matryoshka,devvrit2023matformer}. As with cascade-aware training, one can also aim to train each of the sub-models to be aware of their adaptive usage~\citep{yu2022boostdnn,regol2024jointlylearned}.

An altogether different approach to conditional computation is via the paradigm of `experts as layers'--that is, using a sparse layer consisting of distinct `experts' to effectively replace each dense layer in a model architecture which is otherwise unchanged.  Pursuing conditional computation in this form requires `baking' sparsity into the model at a granular level (i.e., per layer), from the start of training. Notable examples of research on this approach are Mixture-of-Experts ~\citep{shazeer2017moe} and Switch Transformers~\citep{fedus2022switch}). For an excellent overview of this family's history, see~\citet{fedus2022review}, which we will not attempt to reproduce here.

\paragraph{Cascade Deferral Decision-Making}

Cascades involve orchestrating amongst a series of models via a \emph{deferral rule}, which decides which model is most appropriate for a given input.
Classically, for a probabilistic classifier,
this is simply based on thresholding the model's probability for the predicted class, or the entropy of the model's probability distribution~\citep{Viola:2001,Wang:2018}.
While it is possible to \emph{learn} a deferral rule based on these probabilities (and other features)~\citep{Trapeznikov:2013,narasimhan2022posthocest}
---
leveraging advances in the literature on learning to defer to an expert~\citep{Madras:2018,Mozannar:2020,verma2022calibrated,Mao:2023}
---
simple probability thresholding is often competitive~\citep{jitkrittum2023confbaseddef}.
We note also that learned deferral rules can be seen as generalizing \emph{model routing},
wherein a model selector is learned based purely on the input example~\citep{shnitzer2023large,lu2023routing,hari2023tryage,Yiding:2023,lee2024orchestrallm,Sakota:2024,Ding:2024,Hu:2024}.



\paragraph{Data Filtering}
Equation \eqref{eq:cat_loss} can be seen as \emph{filtering} out from the training objective certain `hard-to-learn' tokens, wherein neither the small nor the large model makes a correct prediction. The value of filtering out `hard-to-learn' samples has been demonstrated in conventional classification problems~\citep{2022PrioritizedTraining}, and more broadly, the question of what constitutes an `easy', `hard', or `important' example to learn has been an active thread of research~\citep{toneva2018exforget, ren2018reweightex, paul2021diet, baldock2021exdiff, agarwal2022vog}. A closely related line of work has also demonstrated the value of filtering out samples with noisy labels~\citep{jiang2018mentornet,pmlr-v97-song19b,wei2022self,xia2023combating}.

\section{Conclusion}
\label{sec.conc}

We have presented an approach to modifying the small model in a LM cascade, filtering tokens used in training to make it `aware' of the knowledge of a larger LM situated downstream in the cascade. Experiments with a representative cascade of LLMs, on a variety of fine-tuning datasets of classification and generative tasks, demonstrate the efficacy of this cascade-aware LM training for improving the quality-cost tradeoff curve.

In this work, we have endeavored to understand empirically the relative contributions to cascade improvements from two separate effects: (1) `zero-sum' changes to the knowledge of the small model, as it focuses its capacity on easier topics, and (2) non-`zero-sum' changes, where removing the `hard' examples is tantamount to filtering out label noise in training data. But an even deeper understanding of the two effects would be useful.

We applied our methodology to the most basic form of model cascade (namely, involving just two LMs). A number of natural extensions come to mind. Applying cascade-aware training for a cascade of three or more LMs, where perhaps both the small and medium models have their parameters updated during fine-tuning, would be an interesting study. What should the criteria for filtering tokens (i.e., the $\alpha_i$ term of \eqref{eq:cat_loss}) be in such a scenario?

Another interesting future direction of cascade-aware training would be in regards to the `flavor' of fine-tuning applied. We focused on supervised fine-tuning methods, but reinforcement learning (RL) approaches to fine-tuning, like RLHF, have become of significant interest to the LLM field~\citep{ouyang2022rlhf}.

As noted in the introduction, one of the major benefits of a model cascade is the capacity to host the cascade's models in different physical or logical locations. The techniques we present are naturally amenable to deployment in a distributed system, where e.g.~small models live on edge devices and large models are deployed in a datacenter. In this vein, it would be interesting to explore cascade-aware training further in a setting like federated learning (FL)~\citep{mcmahan2017communication}, where the fine-tuning data is private and decentralized (residing at the edge e.g.~on users' phones), and a global cascade-aware trained small model is learned over successive rounds of FL training. Recent developments in distributed training over edge devices (e.g.~\citet{rush2023federated}) enable development and training against much more flexible losses, including e.g. dynamic system costs. Overall, we expect the exploration of training cascaded models in and for distributed deployments to form a major area of research going forward.

\bibliographystyle{plainnat}
\bibliography{cascade_aware_training}


\clearpage
\newpage

\begin{center}
{\LARGE{}\ourtitle{}}{\LARGE\par}
\par\end{center}

\begin{center}
\textcolor{black}{\Large{}Appendix}{\Large\par}
\par\end{center}

\appendix

\section{Hyperparameters}
\label{app.hyperparam}

In this section we provide details on fine tuning (batch size, which checkpoints, number of training steps, LR, etc) in Table \ref{apdx:tbl:train_config}. For all fine-tuning experiments, we use PaLM-2 \citep{anil2023palm} Gecko and Otter as our small and large LMs, respectively.
We fine-tune them on one of SuperGLUE \citep{wang2019superglue}, WMT22 \citep{kocmi2022wmt} or FLAN2021 \citep{wei2021flan}, before evaluation.
For Gecko fine tuning, we used 16xTPUv4 (HBM2 32GB), with the training time 6-9 hour on SuperGLUE dataset, 2-4 hours on WMT22 dataset and 8-11 hours on FLAN2021 dataset. For Otter fine tuning, we used 128xTPUv5e (HBM2 16GB) with training time roughly 3 days for SuperGLUE dataset and 64xTPUv4 with training time around 2 hours on WMT22 dataset and 3 days on FLAN2021 dataset.
For all fine-tuning tasks or models mentioned in this appendix, they follow the same setting as listed in Table \ref{apdx:tbl:train_config}, unless otherwise indicated. 

\begin{table}[h]
\caption{Training configuration for the LMs tested.}
\vspace{0.2pt}
\begin{tabular}{llllllll}
\hline
Dataset   & Base & Loss        & Train Steps & LR   & Batch Sz & Dropout & Optimizer \\ \hline
SuperGLUE & Gecko      & Xent        & 10000          & 5e-5 & 64         & 0.1          & Adafactor       \\ 
SuperGLUE & Gecko      & CAT-Xent    & 10000          & 5e-5 & 64         & 0.1          & Adafactor       \\ 
SuperGLUE & Gecko      & Dist        & 10000          & 5e-5 & 64         & 0.1          & Adafactor       \\ 
SuperGLUE & Gecko      & CAT-Dist    & 10000          & 5e-5 & 64         & 0.1          & Adafactor    \\ 
SuperGLUE & Otter      & Xent        & 5000           & 1e-5 & 512        & 0.1          & Adafactor       \\ \hline 

WMT22 & Gecko      & Xent        & 4000          & 5e-5 & 32         & 0.1          & Adafactor       \\ 
WMT22 & Gecko      & CAT-Xent    & 4000          & 5e-5 & 32         & 0.1          & Adafactor       \\ 
WMT22 & Gecko      & Dist        & 4000          & 5e-5 & 32         & 0.1          & Adafactor       \\ 
WMT22 & Gecko      & CAT-Dist    & 4000          & 5e-5 & 32         & 0.1          & Adafactor    \\ 
WMT22 & Otter      & Xent        & 1000           & 1e-5 & 1024        & 0.1          & Adafactor       \\ \hline

FLAN2021 & Gecko      & Xent        & 10000          & 1e-3 & 64         & 0.1          & Adafactor       \\ 
FLAN2021 & Gecko      & CAT-Xent    & 10000          & 1e-3 & 64         & 0.1          & Adafactor       \\ 
FLAN2021 & Gecko      & Dist        & 10000          & 1e-3 & 64         & 0.1          & Adafactor       \\ 
FLAN2021 & Gecko      & CAT-Dist    & 10000          & 1e-3 & 64         & 0.1          & Adafactor    \\ 
FLAN2021 & Otter      & Xent        & 60000           & 3e-5 & 128        & 0.05          & Adafactor       \\
\hline
\end{tabular}
\label{apdx:tbl:train_config}
\end{table}

\section{Comparison with other baselines}
\label{app.other_baselines}

\begin{figure}[t]
    \centering
    \begin{minipage}[b]{0.49\textwidth}
        \centering
        \includegraphics[width=\textwidth]{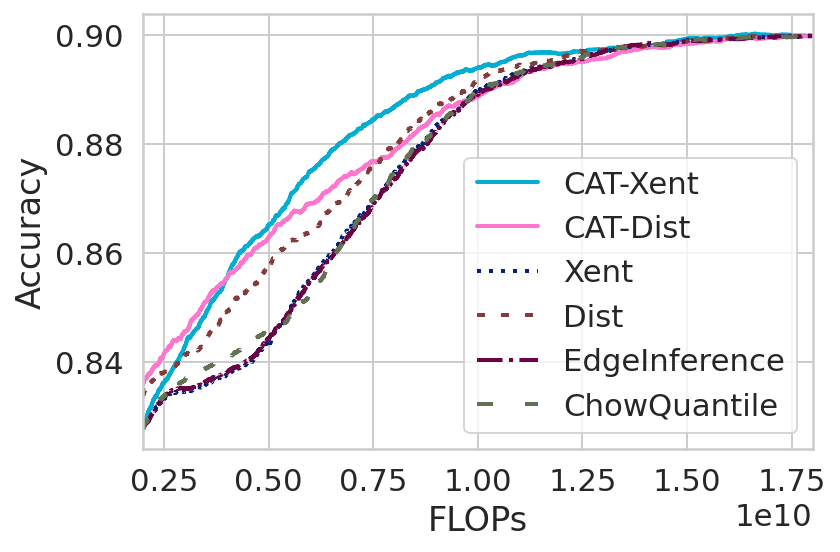}
        \subcaption{SuperGLUE}
    \end{minipage}
    \hfill
    \begin{minipage}[b]{0.49\textwidth}
        \centering
        \includegraphics[width=\textwidth]{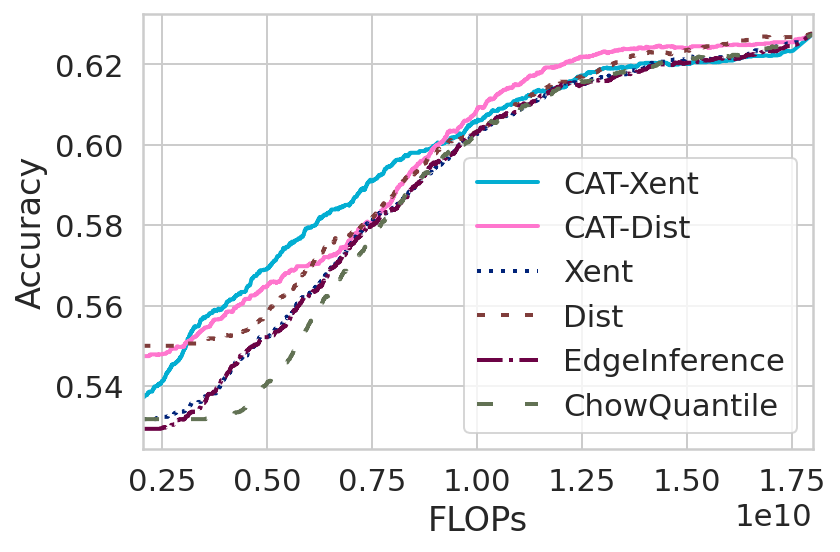}
        \subcaption{FLAN2021-Cls-Tasks}
    \end{minipage}

    \begin{minipage}[b]{0.49\textwidth}
        \centering
        \includegraphics[width=\textwidth]{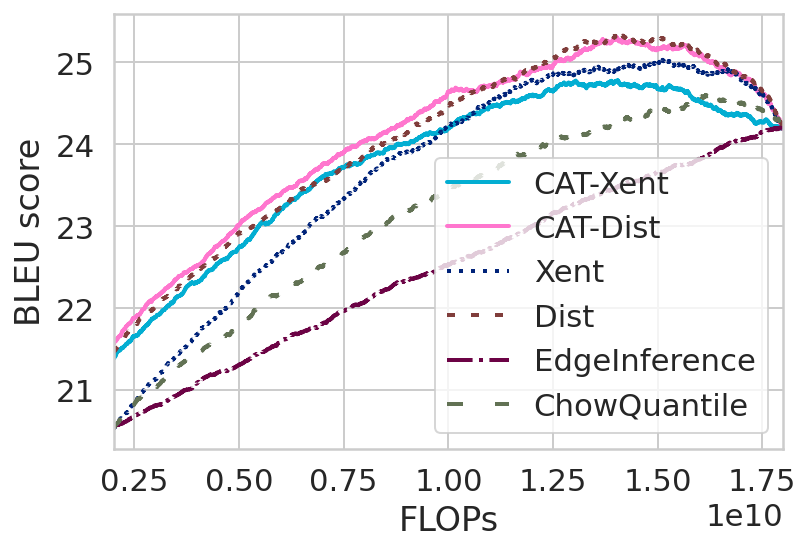}
        \subcaption{WMT22}
    \end{minipage}
    \hfill
    \begin{minipage}[b]{0.49\textwidth}
        \centering
        \includegraphics[width=\textwidth]{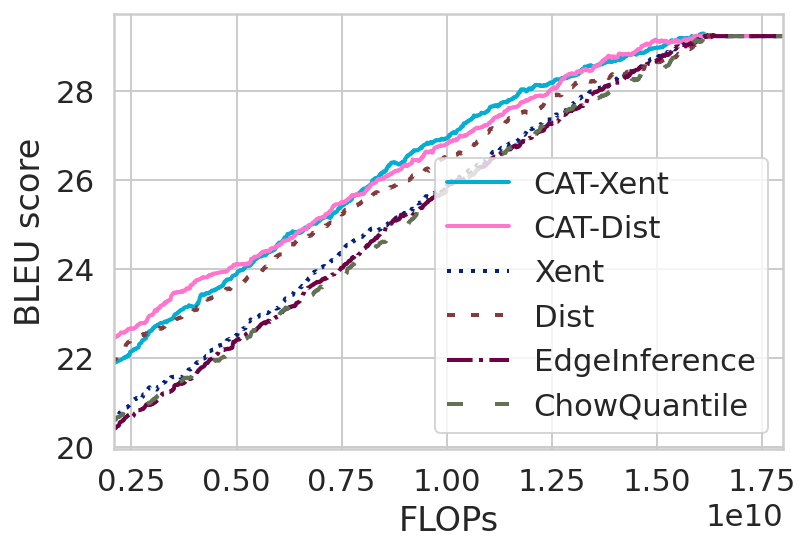}
        \subcaption{FLAN2021-Gen-Tasks}
    \end{minipage}
    \caption{Quality-vs-FLOPs curves for classification and generation tasks on SuperGLUE, WMT22 and FLAN2021 datasets. }
    \label{apdx:fig:main_flops_all_results}
\end{figure}

In addition to the two baselines, {\tt Xent} and {\tt Dist}, mentioned in the paper, we compared CAT against two other baselines: {\tt EdgeInference} \citep{kag2023efficient} and {\tt ChowQuantile} \citep{gupta2024lmcascades}. The {\tt EdgeInference} method, a notable work exploring the benefits of model training in a cascade, uses a router on the output logits or embeddings of the smaller model to determine deferral based on the router's scores. However, this approach was designed for classification tasks in the vision field, where the model generates a single embedding and logits vector. This method is not directly applicable to language model tasks, where outputs are sequences of tokens, whether for classification or generation tasks. To adapt {\tt EdgeInference} for our study, we modified the router’s input from the vector of logits to the logits of each output token from the small language model:

\begin{align}
[p_{S}(y_{S,i}|\mathbf{x},\mathbf{y}_{S,<i})], \quad i \in\{0, 1, 2,...N\},
\label{apdx:eq:routing}
\end{align}

forming a feature vector. Since the output token length is not fixed, we padded it with zeros to a fixed length of 2048. We fine-tuned the router while keeping both the small and large models frozen, with one-hot cross-entropy loss. {\tt ChowQuantile} \citep{gupta2024lmcascades} explores different methods to combine the small model’s logits into reliable deferral indicators. We used {\tt ChowQuantile}-0, which essentially uses the minimum of the logits from the small model as the deferral indicator, as one baseline. In the following section of ablation study, we will present results from other quantile choices.

For {\tt EdgeInference}, its performance is quite similar to the baseline {\tt Xent} on the SuperGLUE dataset and the classification tasks on the FLAN2021 dataset. This is because the response length for classification tasks in language models generally consists of a limited number of tokens (mostly just one token). However, when tested on generation tasks, the cascade quality dropped significantly, especially for the WMT22 dataset, as shown in Figure~\ref{apdx:fig:main_flops_all_results} (c) and (d). Since {\tt ChowQuantile} does not fine-tune the model and relies solely on the model's inherent capability, it uses the same small model as the baseline {\tt Xent}. The results of {\tt ChowQuantile} are generally not on par with {\tt Xent}, which uses the average.


\section{Ablation study}
\label{app.ablation}

In the ablation study, we aim to explore: 1) the best methods to derive deferral decisions from token-level uncertainty, aside from the approach described in \eqref{eq:routing}; and 2) two alternative loss designs that are natural extensions of the current one.

\subsection{Token-level uncertainty to deferral decision}
\begin{figure}[t]
    \centering
    \begin{minipage}[b]{0.49\textwidth}
        \centering
        \includegraphics[width=\textwidth]{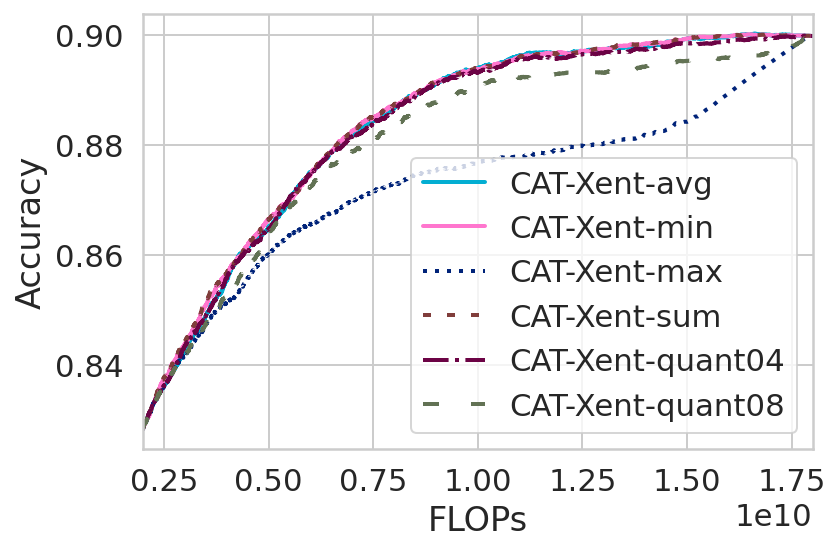}
        \subcaption{SuperGLUE with CAT-Xent loss}
    \end{minipage}
    \hfill
    \begin{minipage}[b]{0.49\textwidth}
        \centering
        \includegraphics[width=\textwidth]{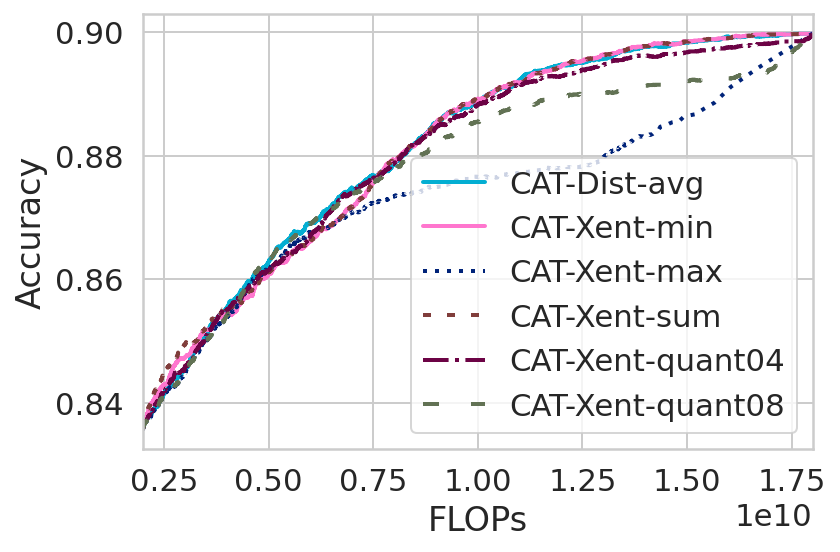}
        \subcaption{SuperGLUE with CAT-Dist loss}
    \end{minipage}

    \begin{minipage}[b]{0.49\textwidth}
        \centering
        \includegraphics[width=\textwidth]{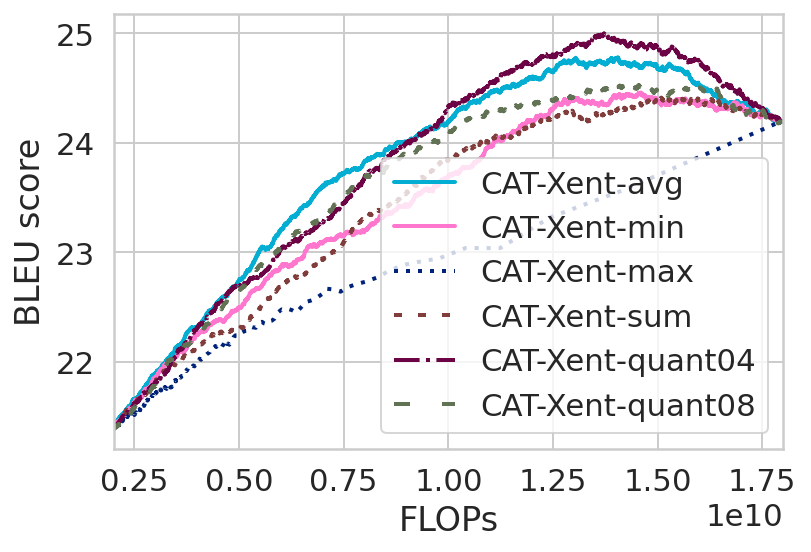}
        \subcaption{WMT22 with CAT-Xent loss}
    \end{minipage}
    \hfill
    \begin{minipage}[b]{0.49\textwidth}
        \centering
        \includegraphics[width=\textwidth]{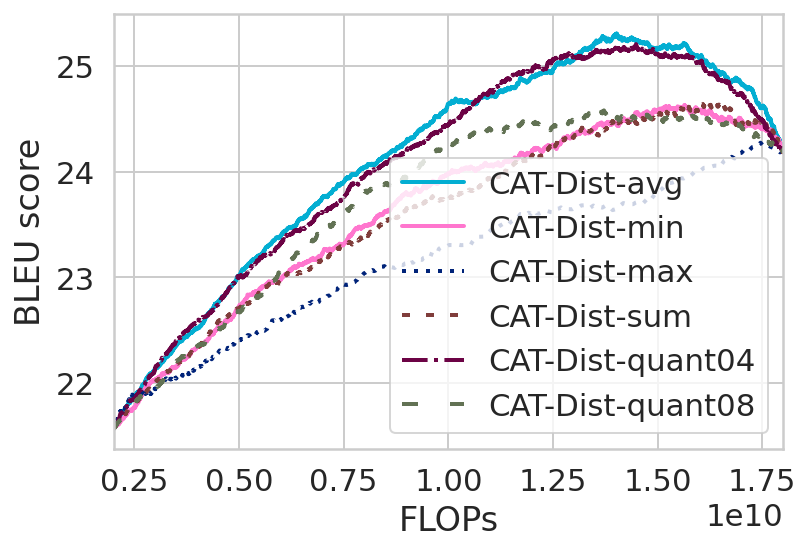}
        \subcaption{WMT22 with CAT-Dist loss}
    \end{minipage}
    \caption{Quality-vs-FLOPs curves for different deferral rules on SuperGLUE, and WMT22 datasets. }
    \label{apdx:fig:deferral_rules}
\end{figure}

This exploration is similar to \citet{gupta2024lmcascades}. The major difference is that we are exploring deferral rules in a multi-task scenario, whereas \citet{gupta2024lmcascades} focused on single-task deferral.

As indicated in \eqref{eq:routing}, we can use token-wise logits from the language model to measure its uncertainty (or confidence) in its responses. We compared six different methods to combine logits from tokens into a deferral indicator: 'average', 'minimum', 'maximum', 'sum', 'ChowQuantile04', and 'ChowQuantile08' \citep{gupta2024lmcascades}. Note that 'minimum' can be viewed as a special case of 'ChowQuantile' as mentioned in \citep{gupta2024lmcascades}. We tested each deferral rule with our small language model trained with CAT-Xent loss and CAT-Dist loss (\ref{sec.cat_algos}) on the SuperGLUE and WMT22 datasets.

From Figure~\ref{apdx:fig:deferral_rules}, we can clearly see that 'average' works best for both classification and generation tasks with both CAT-Xent and CAT-Dist loss designs. Similar to \citep{gupta2024lmcascades}, we observe that 'ChowQuantile04' (CAT-Xent-quant04 and CAT-Dist-quant04) can be another comparable choice for building the deferral indicator.

\subsection{Four other CAT loss designs}
\label{apdx:sec.ablation_loss_design}
As shown in \eqref{eq:cat_loss}, our loss function essentially filters out tokens that cannot be correctly predicted by either the small or large language model. It is natural to question whether it is sufficient to use only the small or large language model for this filtering. To investigate this, we conducted an ablation study on the SuperGLUE and WMT22 datasets.

In this study, in addition to the {\tt CAT-Xent} and {\tt CAT-Dist} loss designs presented in the main paper, we included four other loss designs.

\textbf{CAT-Xent-L:}
\begin{align}
L_{\mathrm{cat-xent-l}}(\mathbf{x},\mathbf{y}) & =-\sum_{i=1}^{N}\alpha_{i}  \cdot \log p_{S}(y_{i}|\mathbf{x},\mathbf{y}_{<i}) \nonumber \\
\alpha_{i} & :=1\left[y_i=\arg\max_{y' \in \mathcal{V}}p_{L}(y'|\mathbf{x},\mathbf{y}_{<i})\right]
\end{align}

\textbf{CAT-Xent-S:}
\begin{align}
L_{\mathrm{cat-xent-s}}(\mathbf{x},\mathbf{y}) & =-\sum_{i=1}^{N}\alpha_{i}  \cdot \log p_{S}(y_{i}|\mathbf{x},\mathbf{y}_{<i}) \nonumber \\
\alpha_{i} & :=1\left[y_i=\arg\max_{y' \in \mathcal{V}}p_{S}(y'|\mathbf{x},\mathbf{y}_{<i})\right]
\end{align}

\textbf{CAT-Dist-L:}
\begin{align}
L_{\mathrm{cat-dist-l}}(\mathbf{x},\mathbf{y}) & =-\sum_{i=1}^{N}\alpha_{i} \Bigl(w \cdot \log p_{S}(y_{i}|\mathbf{x},\mathbf{y}_{<i}) + (1-w) \cdot \sum_{y'=1}^{V}p_{L}(y'|\mathbf{x},\mathbf{y}_{<i})\log p(y'|\mathbf{x},\mathbf{y}_{<i})\nonumber\Bigr) \\
\alpha_{i} & :=1\left[y_i=\arg\max_{y' \in \mathcal{V}}p_{L}(y'|\mathbf{x},\mathbf{y}_{<i})\right]
\end{align}

\begin{figure}[t]
    \centering
    \begin{minipage}[b]{0.49\textwidth}
        \centering
        \includegraphics[width=\textwidth]{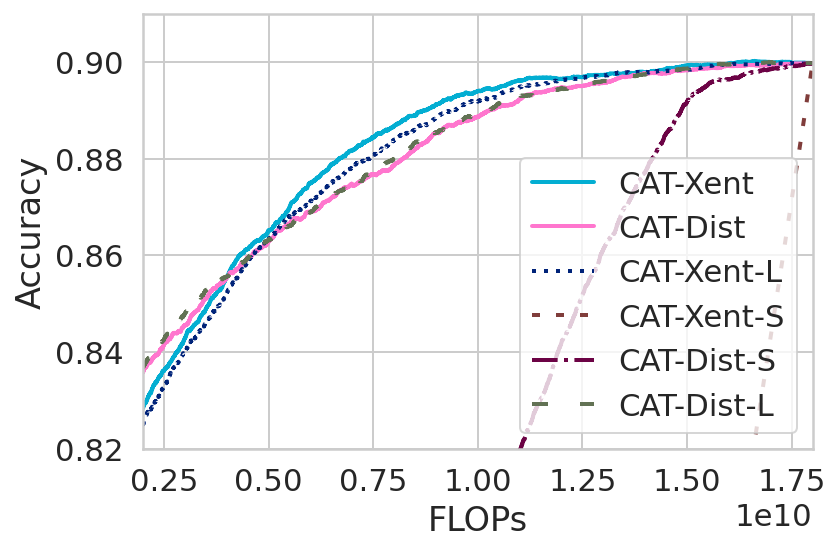}
        \subcaption{Different loss designs on SuperGLUE}
    \end{minipage}
    \hfill
    \begin{minipage}[b]{0.49\textwidth}
        \centering
        \includegraphics[width=\textwidth]{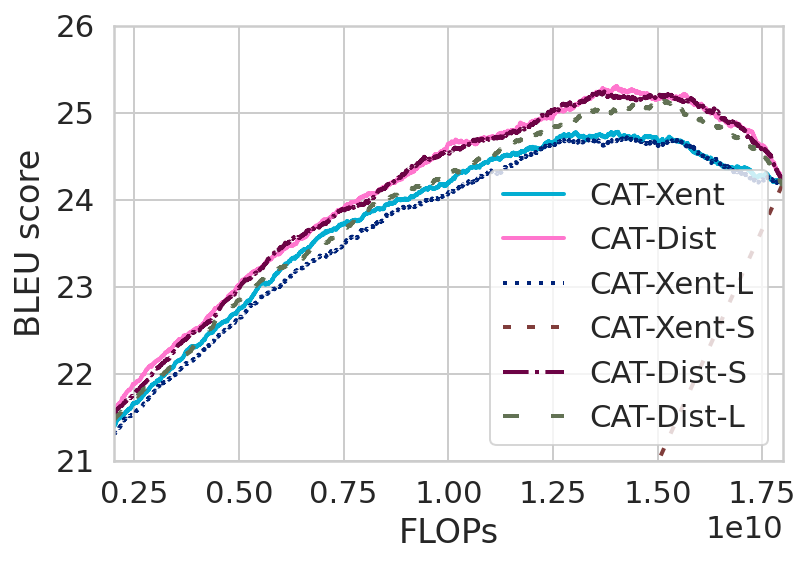}
        \subcaption{Different loss designs on WMT22}
    \end{minipage}
\caption{Ablation study on different cascade-aware loss designs.}
    \label{apdx:fig:loss_design}
\end{figure}

\textbf{CAT-Dist-S:}
\begin{align}
L_{\mathrm{cat-dist-s}}(\mathbf{x},\mathbf{y}) & =-\sum_{i=1}^{N}\alpha_{i} \Bigl(w \cdot \log p_{S}(y_{i}|\mathbf{x},\mathbf{y}_{<i}) + (1-w) \cdot \sum_{y'=1}^{V}p_{L}(y'|\mathbf{x},\mathbf{y}_{<i})\log p(y'|\mathbf{x},\mathbf{y}_{<i})\nonumber\Bigr) \\
\alpha_{i} & :=1\left[y_i=\arg\max_{y' \in \mathcal{V}}p_{S}(y'|\mathbf{x},\mathbf{y}_{<i})\right]
\end{align}
From the results shown in Figure~ \ref{apdx:fig:loss_design}, we found that {\tt CAT-Xent-L} and {\tt CAT-Dist-L} both perform quite competitively compared to our loss design in the main paper. This is reasonable because the large language model should dominate the token filtering with higher accuracy, especially in the early stages of training. Beyond the tokens that the large language model can correctly predict, our loss design in \eqref{eq:cat_loss} also includes those that the small language model can correctly predict. This allows the small language model some extra space to explore, thus achieving the best overall performance compared to other choices.

We also found that training is very slow if we filter out all the tokens that the small language model cannot correctly predict ({\tt CAT-Xent-S} and {\tt CAT-Dist-S}) as shown in Figure~ \ref{apdx:fig:loss_design}. With the same training time, the model will be far from convergence, leading to downgraded cascade performance.

\begin{figure}[t]
    \centering
    \begin{minipage}[b]{0.24\textwidth}
        \centering
        \includegraphics[width=\textwidth]{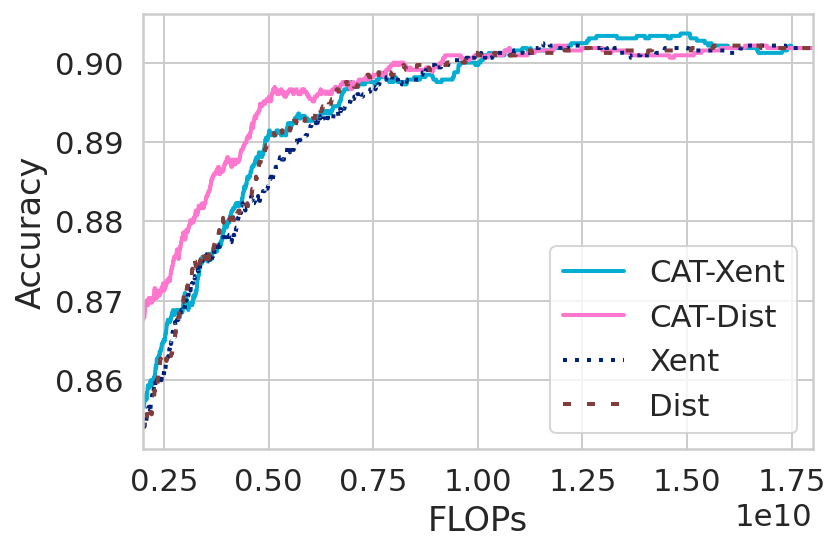}
        \subcaption{BoolQ (3270)}
    \end{minipage}
    \hfill
    \begin{minipage}[b]{0.24\textwidth}
        \centering
        \includegraphics[width=\textwidth]{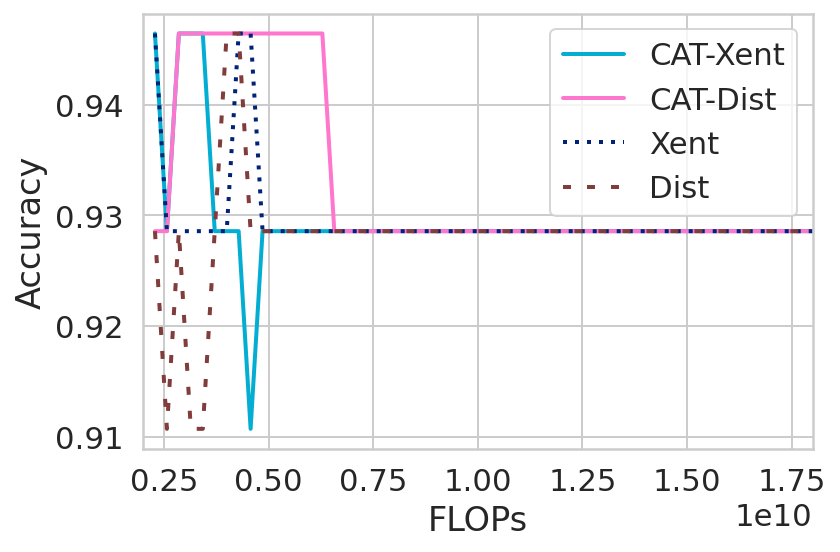}
        \subcaption{CB (56)}
    \end{minipage}
    \hfill
    \begin{minipage}[b]{0.24\textwidth}
        \centering
        \includegraphics[width=\textwidth]{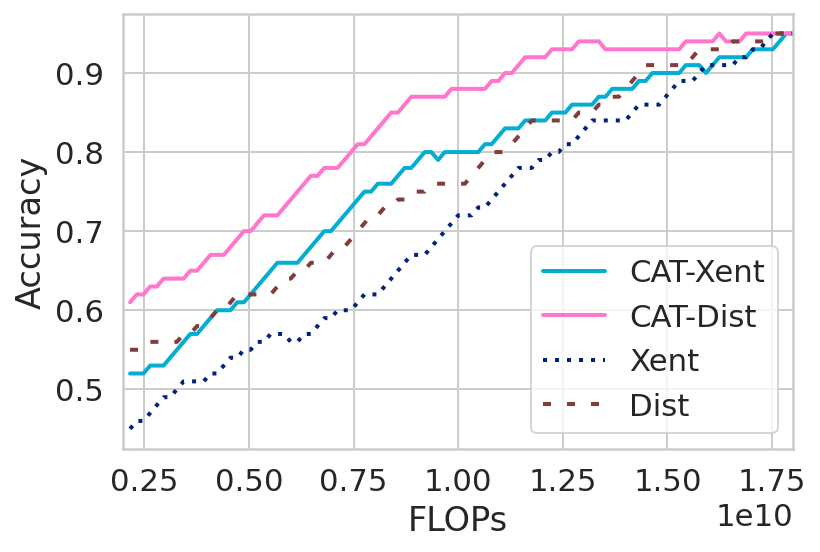}
        \subcaption{Copa (100)}
    \end{minipage}
    \hfill
    \begin{minipage}[b]{0.24\textwidth}
        \centering
        \includegraphics[width=\textwidth]{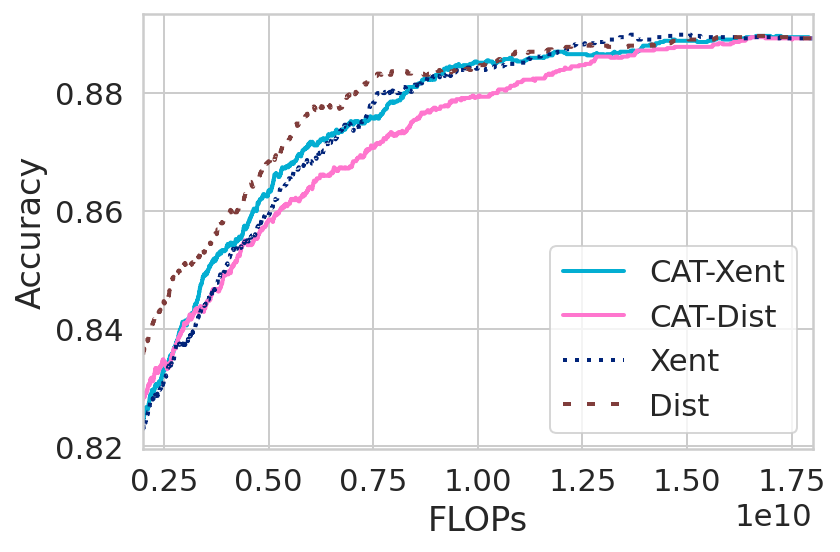}
        \subcaption{MultiRC (4847)}
    \end{minipage}
    \hfill
    \begin{minipage}[b]{0.24\textwidth}
        \centering
        \includegraphics[width=\textwidth]{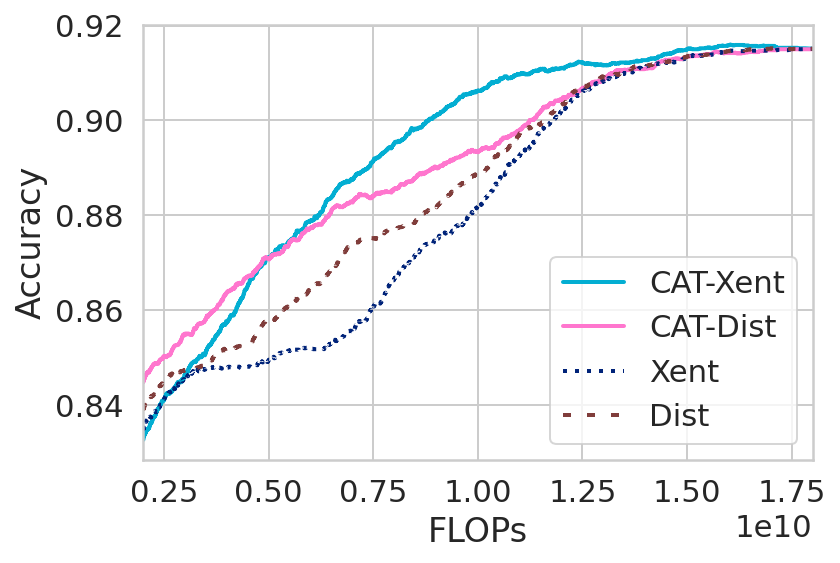}
        \subcaption{Record (10000)}
    \end{minipage}
    \hfill
    \begin{minipage}[b]{0.24\textwidth}
        \centering
        \includegraphics[width=\textwidth]{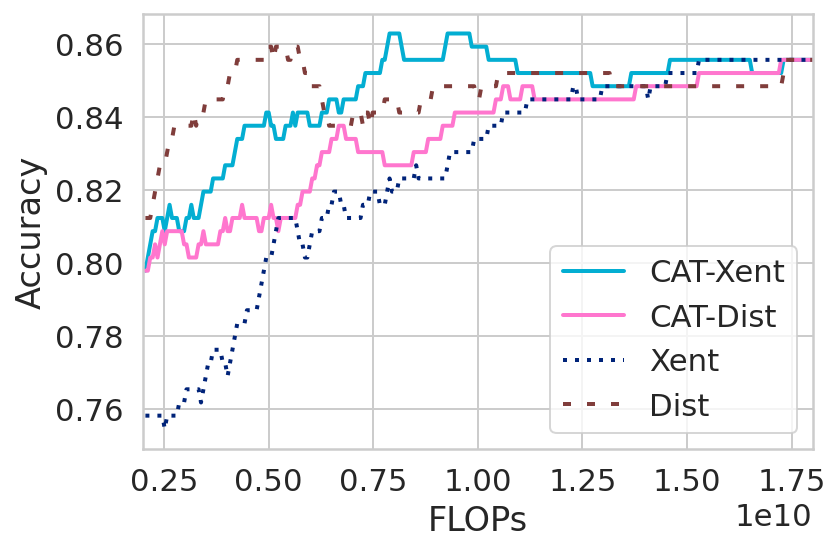}
        \subcaption{RTE (277)}
    \end{minipage}
    \hfill
    \begin{minipage}[b]{0.24\textwidth}
        \centering
        \includegraphics[width=\textwidth]{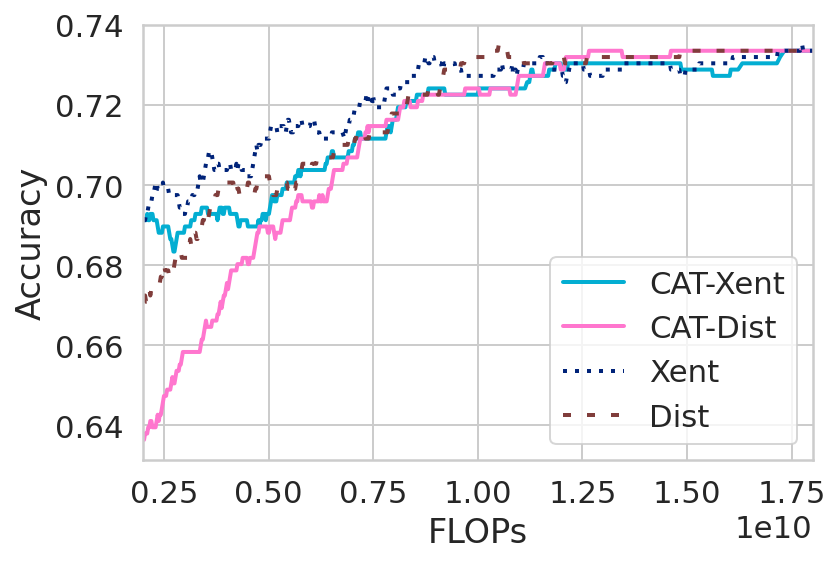}
        \subcaption{WIC (638)}
    \end{minipage}
    \hfill
    \begin{minipage}[b]{0.24\textwidth}
        \centering
        \includegraphics[width=\textwidth]{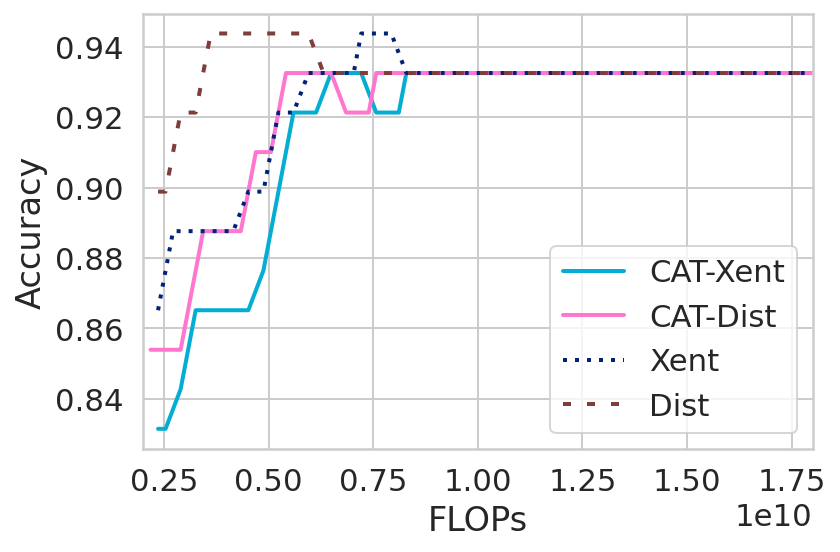}
        \subcaption{WSC (89)}
    \end{minipage}
\caption{Task-specific Quality-FLOPs tradeoff on SuperGLUE dataset.}
    \label{apdx:fig:sg_task_specific}
\end{figure}

\section{Task-specific Quality-FLOPs tradeoff}
\label{app.task_specific}

In the main paper, we presented the results of CAT on three datasets. For each dataset, we combined all tasks, which is more realistic for multi-task LM deployment. In this section, we include the Quality-FLOPs tradeoff in a task-specific manner to provide a better understanding of CAT for each task.

\begin{figure}[t]
    \centering
    \begin{minipage}[b]{0.32\textwidth}
        \centering
        \includegraphics[width=\textwidth]{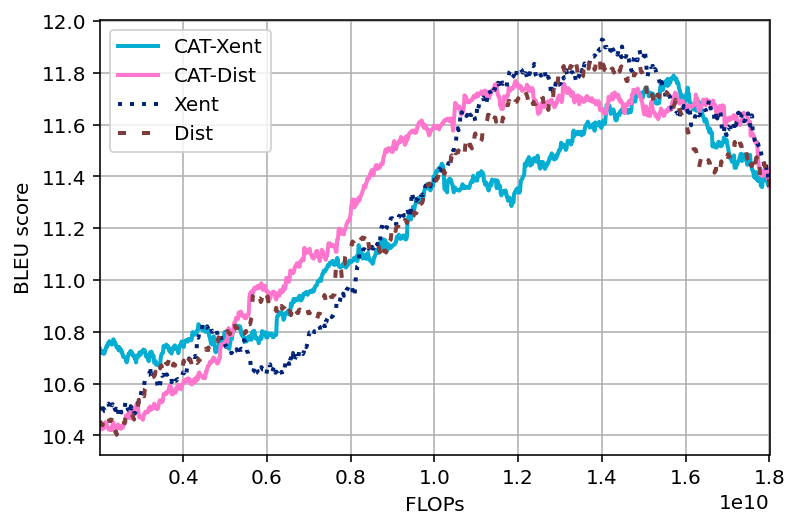}
        \subcaption{En$\rightarrow$Ja (1000)}
    \end{minipage}
    \hfill
    \begin{minipage}[b]{0.32\textwidth}
        \centering
        \includegraphics[width=\textwidth]{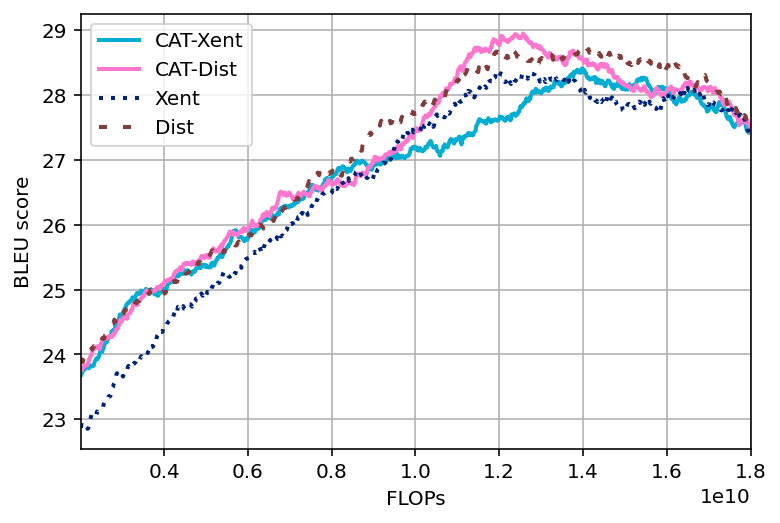}
        \subcaption{En$\rightarrow$Ru (1002)}
    \end{minipage}
    \hfill
    \begin{minipage}[b]{0.32\textwidth}
        \centering
        \includegraphics[width=\textwidth]{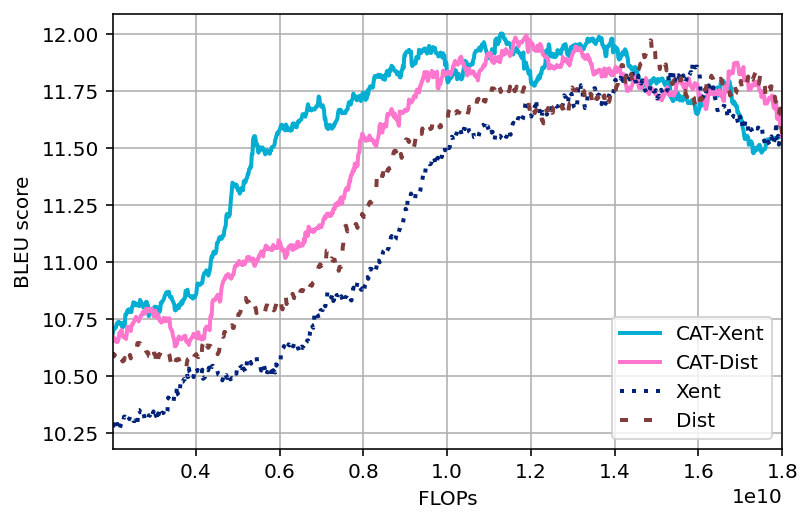}
        \subcaption{En$\rightarrow$Zh (1002)}
    \end{minipage}
    \hfill
    \begin{minipage}[b]{0.32\textwidth}
        \centering
        \includegraphics[width=\textwidth]{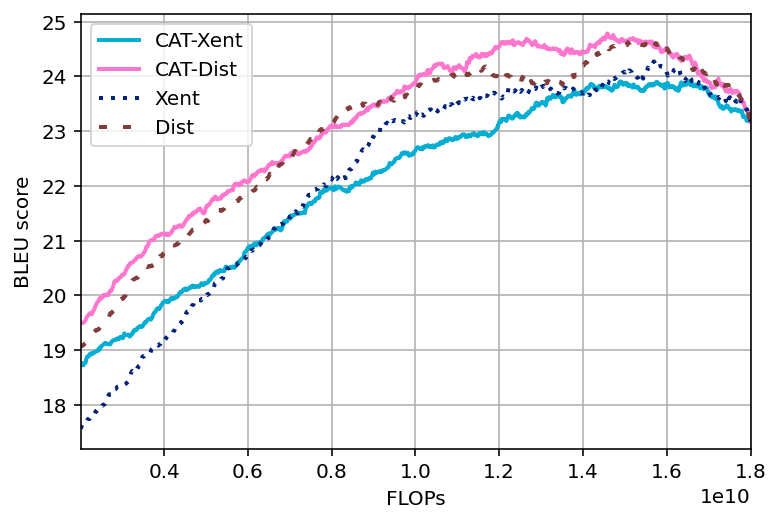}
        \subcaption{Ja$\rightarrow$En (1005)}
    \end{minipage}
    \hfill
    \begin{minipage}[b]{0.32\textwidth}
        \centering
        \includegraphics[width=\textwidth]{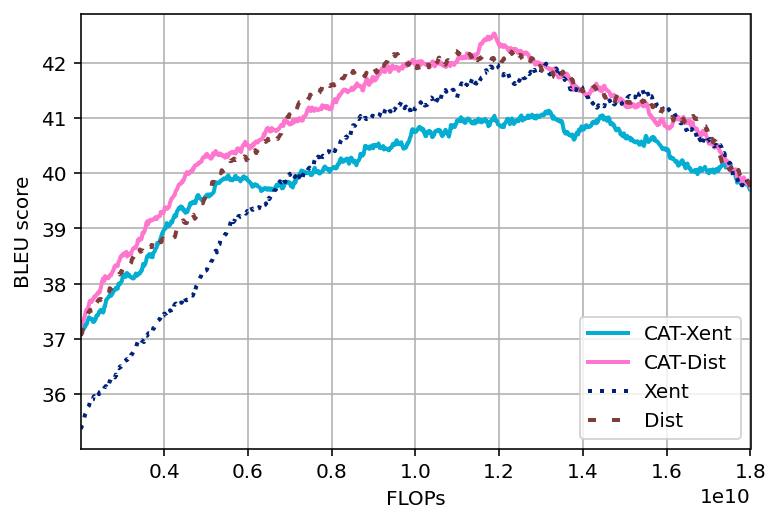}
        \subcaption{Ru$\rightarrow$En (1000)}
    \end{minipage}
    \hfill
    \begin{minipage}[b]{0.32\textwidth}
        \centering
        \includegraphics[width=\textwidth]{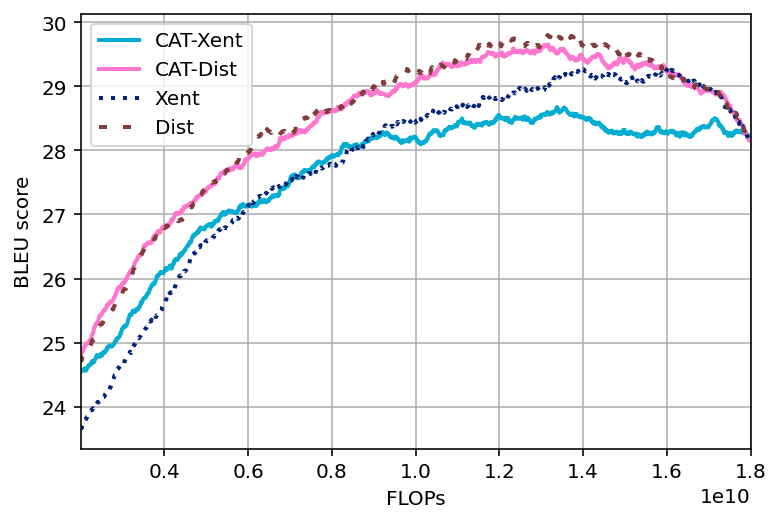}
        \subcaption{Zh$\rightarrow$En (1948)}
    \end{minipage}
\caption{Task-specific Quality-FLOPs tradeoff on WMT22 dataset.}
    \label{apdx:fig:wmt_task_specific}
\end{figure}

\begin{figure}[t]
    \centering
    \begin{minipage}[b]{0.24\textwidth}
        \centering
        \includegraphics[width=\textwidth]{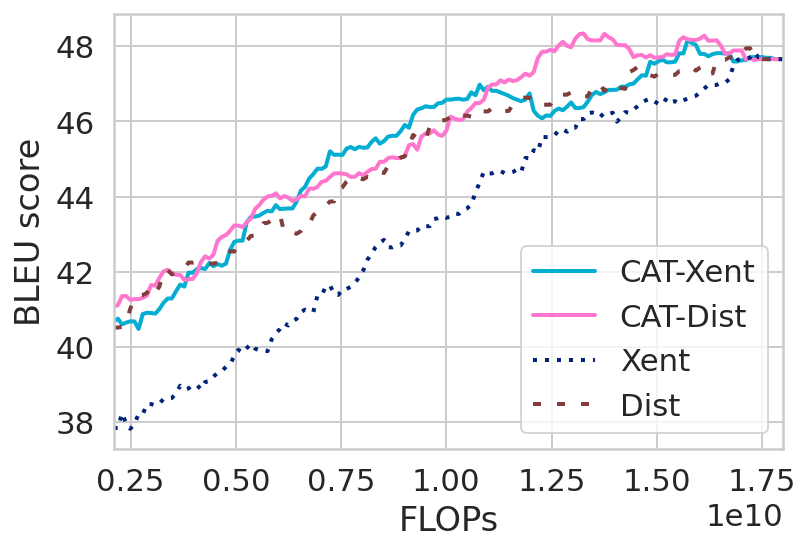}
        \subcaption{ParaCrawl-enes (162)}
    \end{minipage}
    \hfill
    \begin{minipage}[b]{0.24\textwidth}
        \centering
        \includegraphics[width=\textwidth]{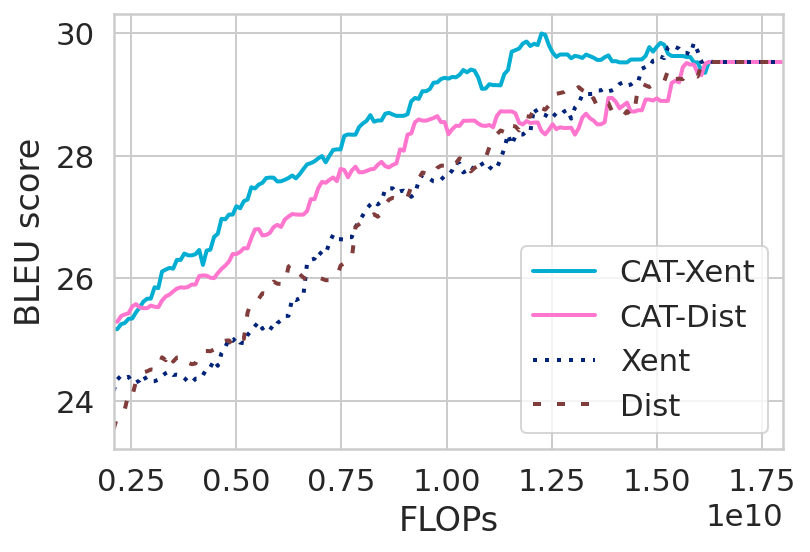}
        \subcaption{WMT14-enfr (181)}
    \end{minipage}
    \hfill
    \begin{minipage}[b]{0.24\textwidth}
        \centering
        \includegraphics[width=\textwidth]{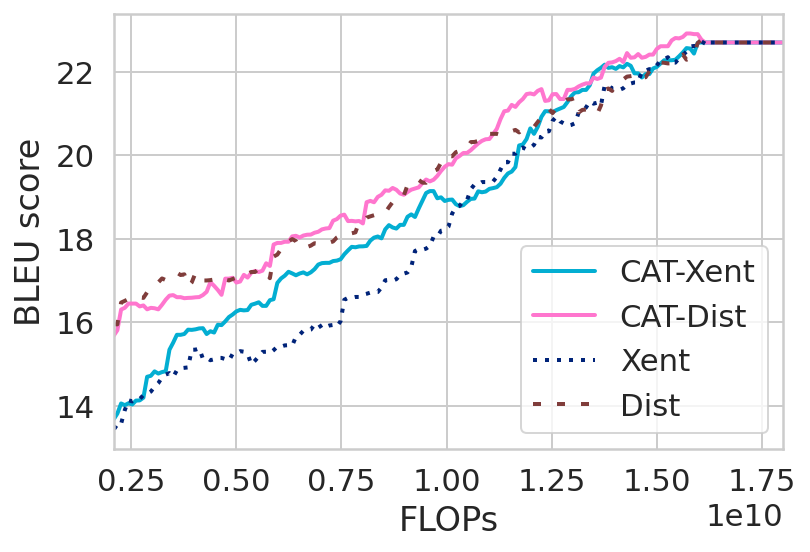}
        \subcaption{WMT16-csen (181)}
    \end{minipage}
    \hfill
    \begin{minipage}[b]{0.24\textwidth}
        \centering
        \includegraphics[width=\textwidth]{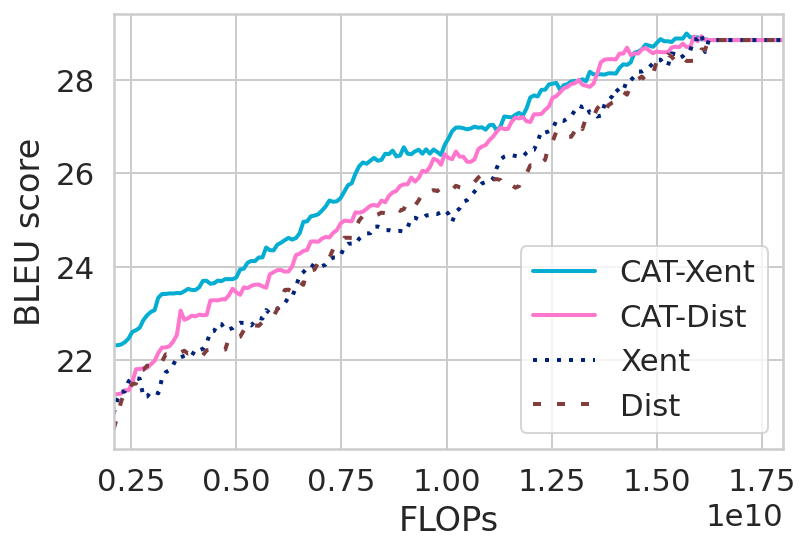}
        \subcaption{WMT16-deen (181)}
    \end{minipage}
    \hfill
    \begin{minipage}[b]{0.24\textwidth}
        \centering
        \includegraphics[width=\textwidth]{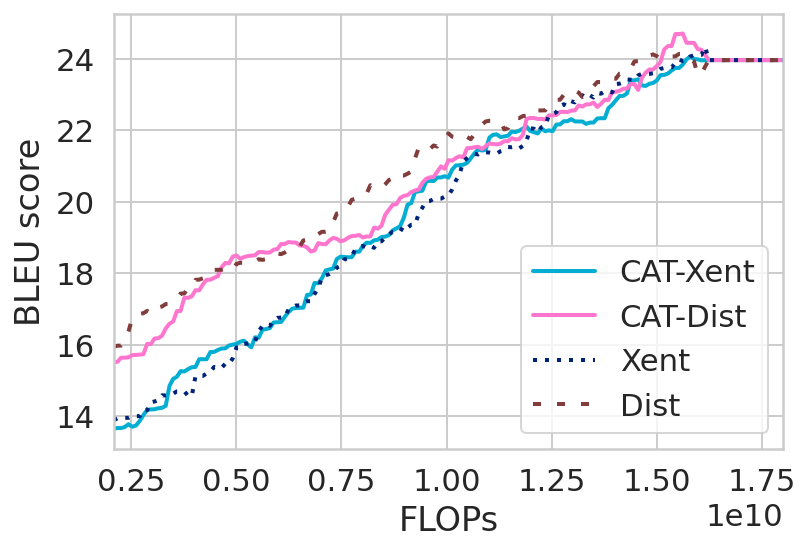}
        \subcaption{WMT16-fien (181)}
    \end{minipage}
    \hfill
    \begin{minipage}[b]{0.24\textwidth}
        \centering
        \includegraphics[width=\textwidth]{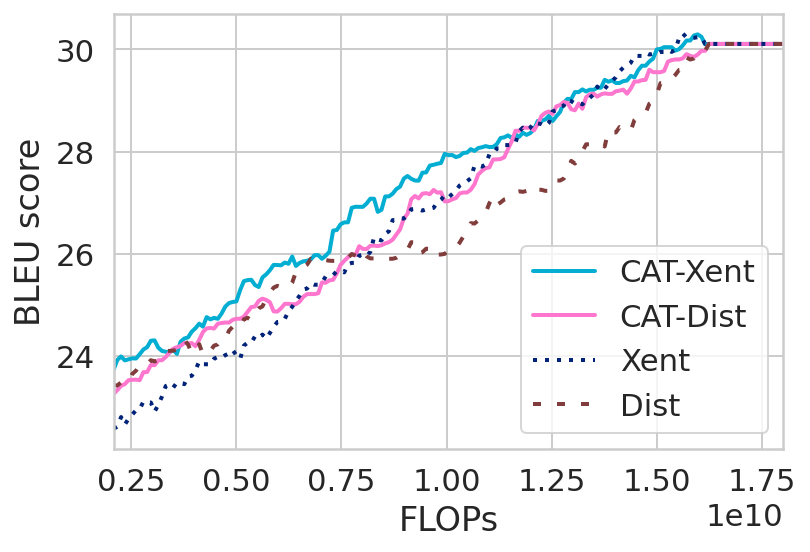}
        \subcaption{WMT16-roen (181)}
    \end{minipage}
    \hfill
    \begin{minipage}[b]{0.24\textwidth}
        \centering
        \includegraphics[width=\textwidth]{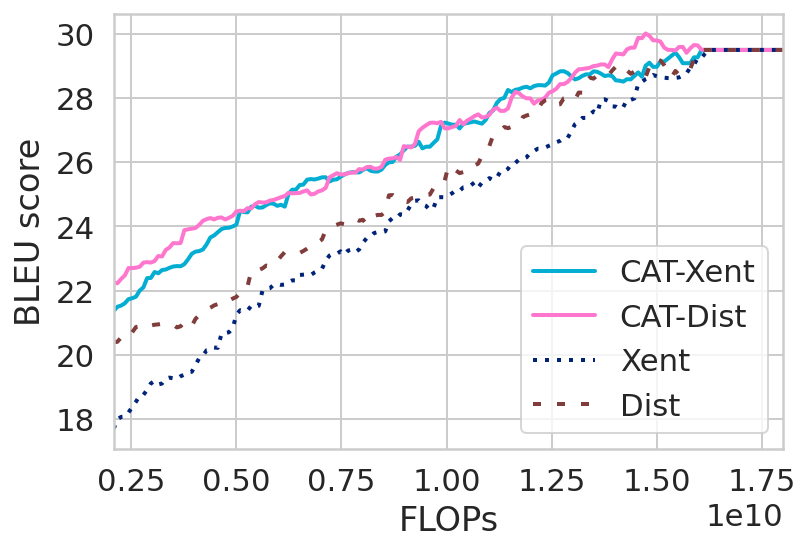}
        \subcaption{WMT16-ruen (181)}
    \end{minipage}
    \hfill
    \begin{minipage}[b]{0.24\textwidth}
        \centering
        \includegraphics[width=\textwidth]{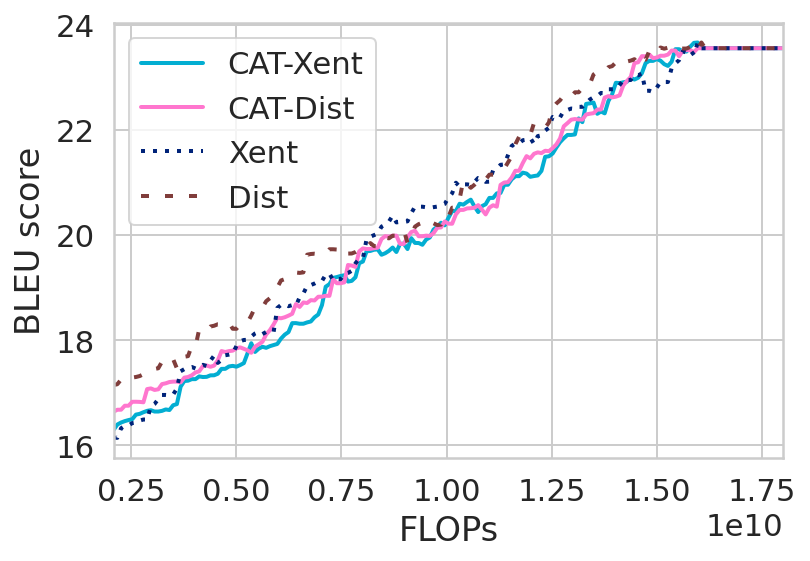}
        \subcaption{WMT16-tren (181)}
    \end{minipage}
\caption{Task-specific Quality-FLOPs tradeoff on FLAN dataset with BLEU score as quality measurement.}
    \label{apdx:fig:flan_bleu_task_specific}
\end{figure}

\begin{figure}[h]
    \centering
    \renewcommand{\thesubfigure}{\arabic{subfigure}}
    \begin{minipage}[b]{0.24\textwidth}
        \centering
        \includegraphics[width=\textwidth]{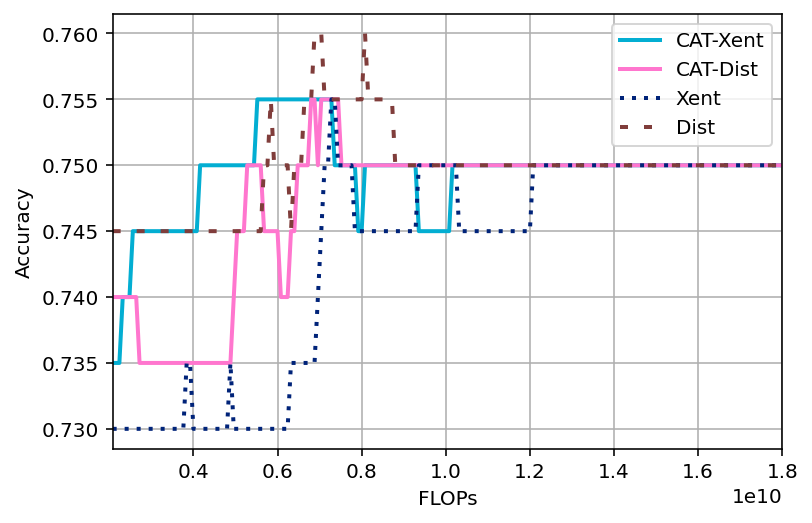}
        \subcaption{Ag News (200)}
    \end{minipage}
    \hfill
    \begin{minipage}[b]{0.24\textwidth}
        \centering
        \includegraphics[width=\textwidth]{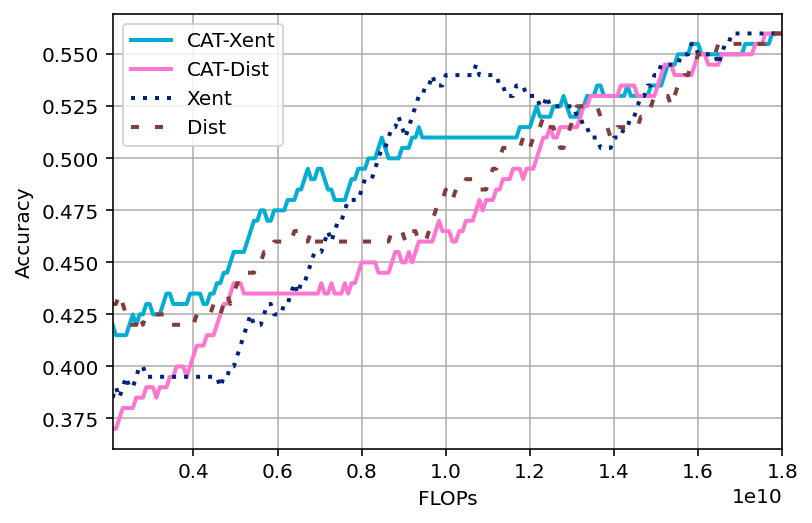}
        \subcaption{Anli (200)}
    \end{minipage}
    \hfill
    \begin{minipage}[b]{0.24\textwidth}
        \centering
        \includegraphics[width=\textwidth]{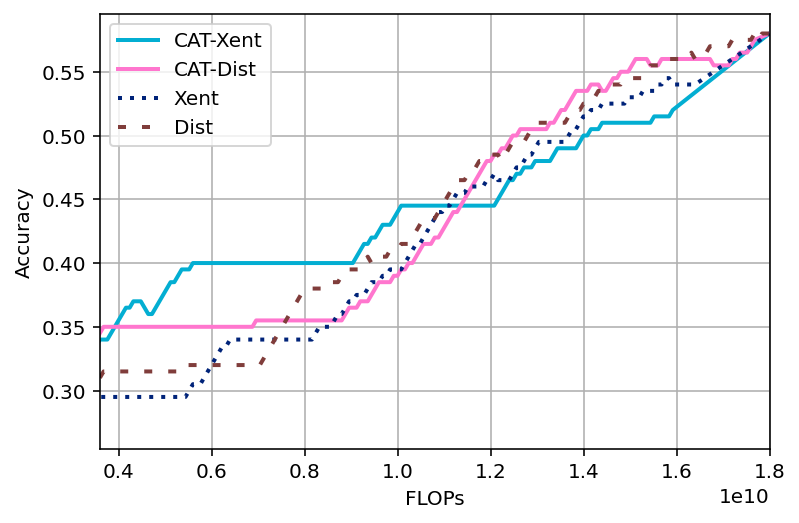}
        \subcaption{Arc-challenge (200)}
    \end{minipage}
    \hfill
    \begin{minipage}[b]{0.24\textwidth}
        \centering
        \includegraphics[width=\textwidth]{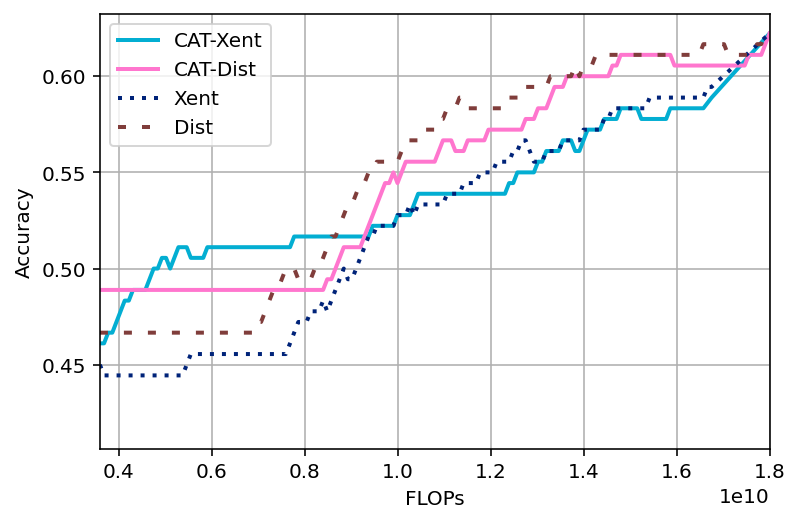}
        \subcaption{Arc-easy (180)}
    \end{minipage}
    \hfill
    \begin{minipage}[b]{0.24\textwidth}
        \centering
        \includegraphics[width=\textwidth]{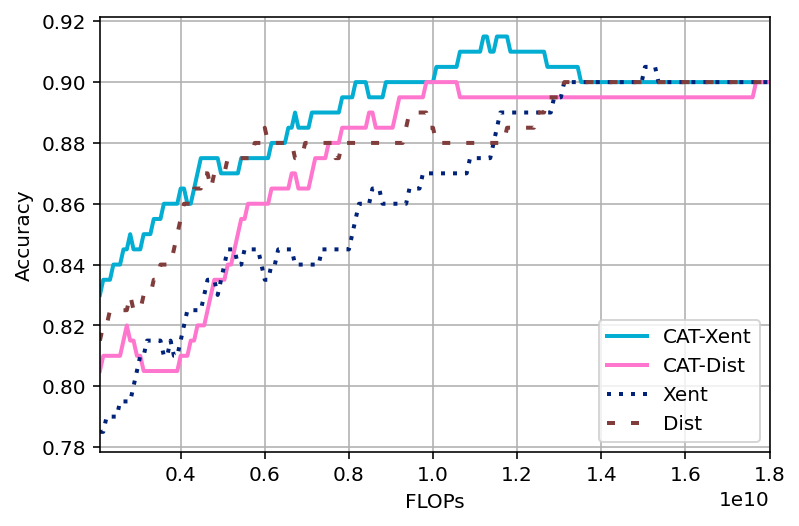}
        \subcaption{BoolQ (200)}
    \end{minipage}
    \hfill
    \begin{minipage}[b]{0.24\textwidth}
        \centering
        \includegraphics[width=\textwidth]{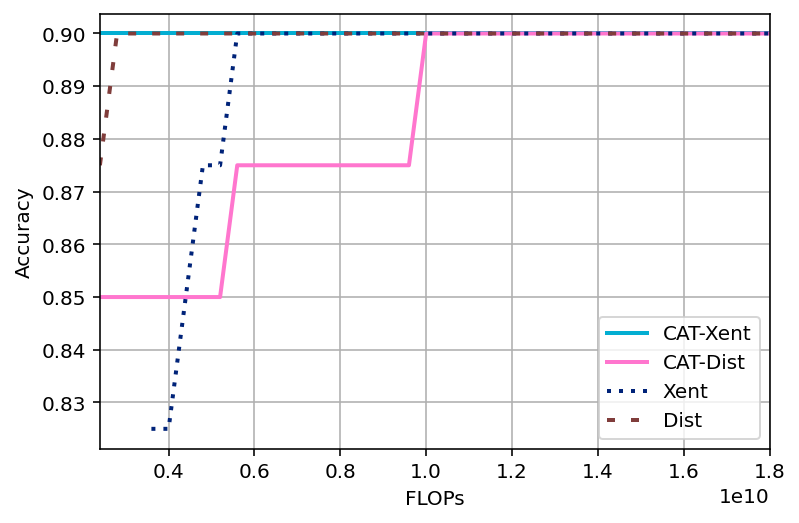}
        \subcaption{CB (40)}
    \end{minipage}
    \hfill
    \begin{minipage}[b]{0.24\textwidth}
        \centering
        \includegraphics[width=\textwidth]{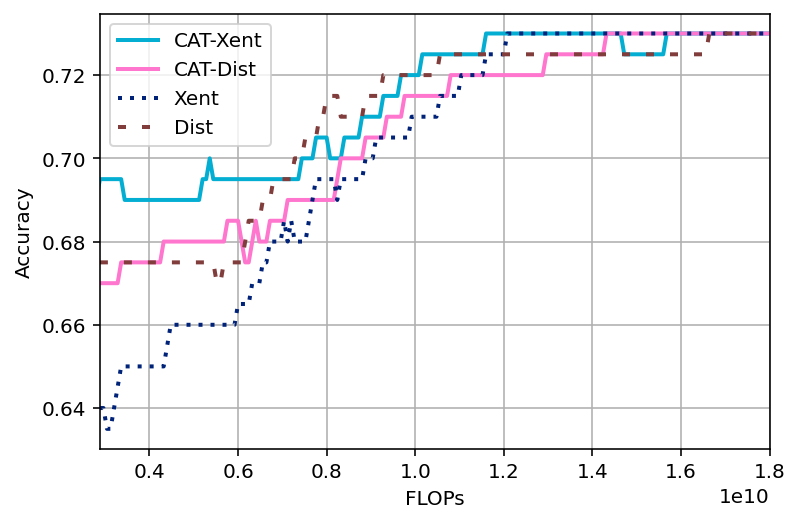}
        \subcaption{Cola (200)}
    \end{minipage}
    \hfill
    \begin{minipage}[b]{0.24\textwidth}
        \centering
        \includegraphics[width=\textwidth]{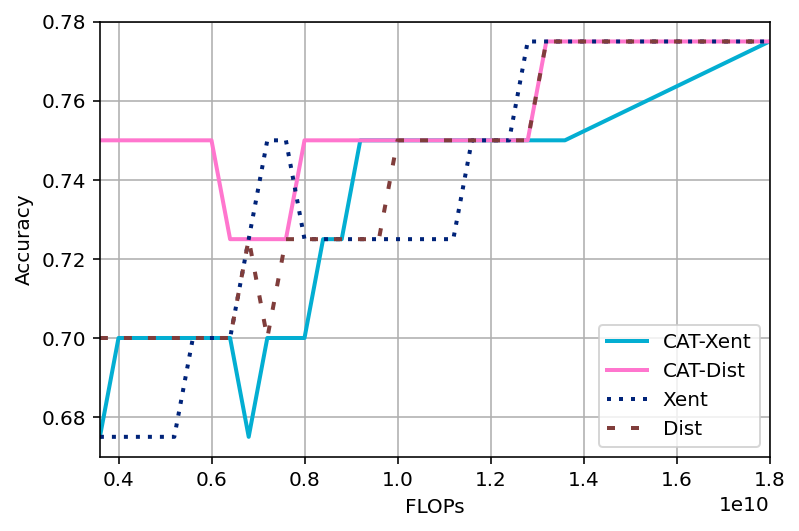}
        \subcaption{Copa (40)}
    \end{minipage}
    \hfill
    \begin{minipage}[b]{0.24\textwidth}
        \centering
        \includegraphics[width=\textwidth]{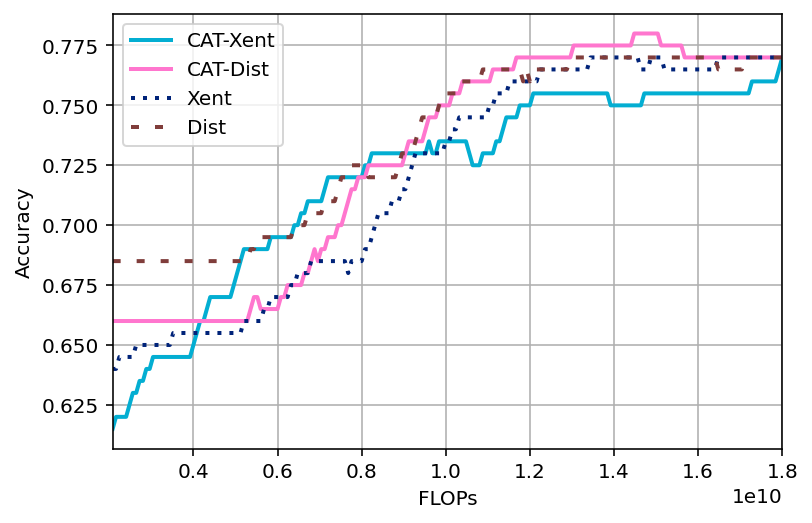}
        \subcaption{Cosmos QA (200)}
    \end{minipage}
    \hfill
    \begin{minipage}[b]{0.24\textwidth}
        \centering
        \includegraphics[width=\textwidth]{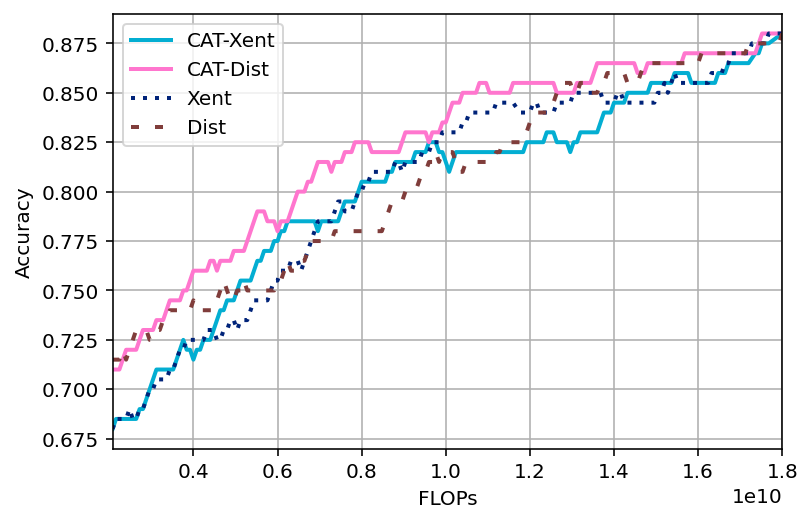}
        \subcaption{Definite-pronoun-res (200)}
    \end{minipage}
    \hfill
    \begin{minipage}[b]{0.24\textwidth}
        \centering
        \includegraphics[width=\textwidth]{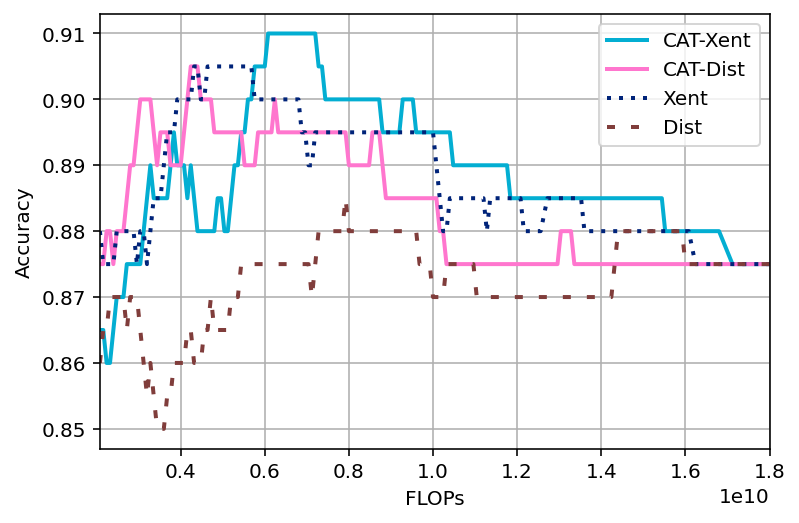}
        \subcaption{Glue mrpc (200)}
    \end{minipage}
    \hfill
    \begin{minipage}[b]{0.24\textwidth}
        \centering
        \includegraphics[width=\textwidth]{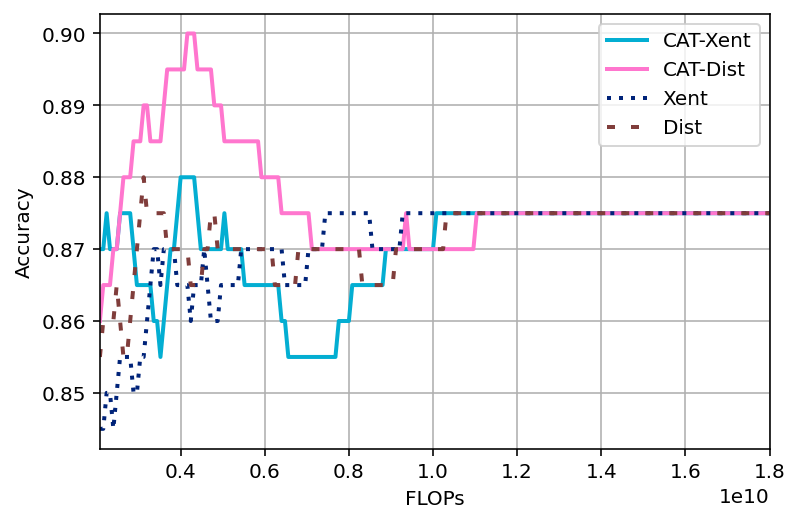}
        \subcaption{Glue qqp (200)}
    \end{minipage}
    \hfill
    \begin{minipage}[b]{0.24\textwidth}
        \centering
        \includegraphics[width=\textwidth]{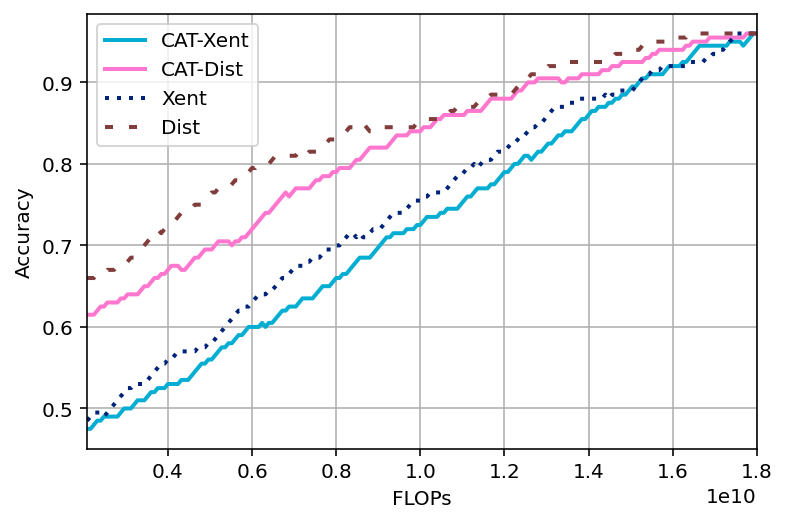}
        \subcaption{HellaSwag (200)}
    \end{minipage}
    \hfill
    \begin{minipage}[b]{0.24\textwidth}
        \centering
        \includegraphics[width=\textwidth]{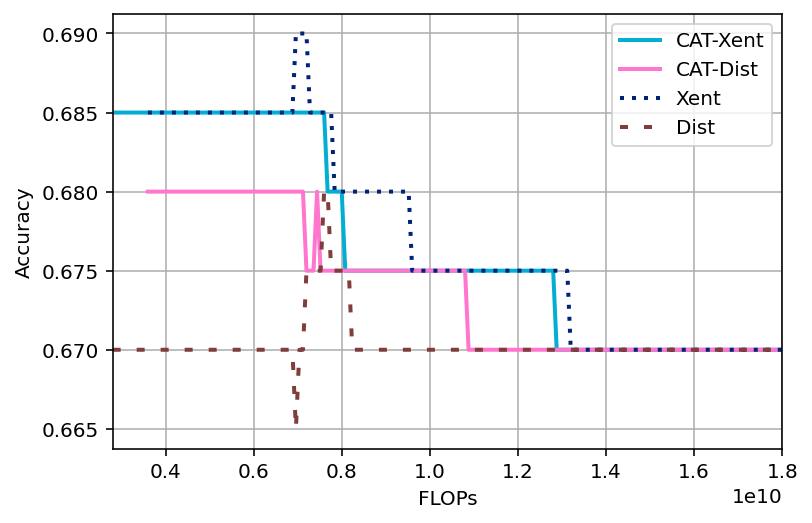}
        \subcaption{Imdb reviews (200)}
    \end{minipage}
    \hfill
    \begin{minipage}[b]{0.24\textwidth}
        \centering
        \includegraphics[width=\textwidth]{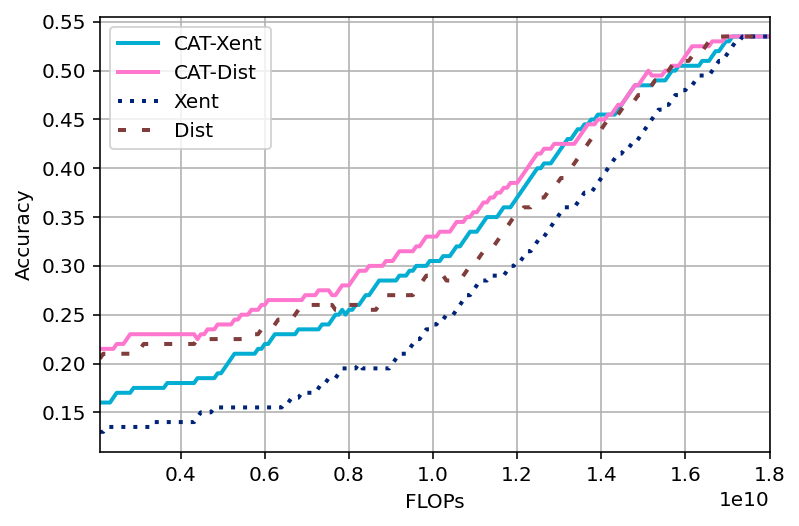}
        \subcaption{Math Dataset (200)}
    \end{minipage}
    \hfill
    \begin{minipage}[b]{0.24\textwidth}
        \centering
        \includegraphics[width=\textwidth]{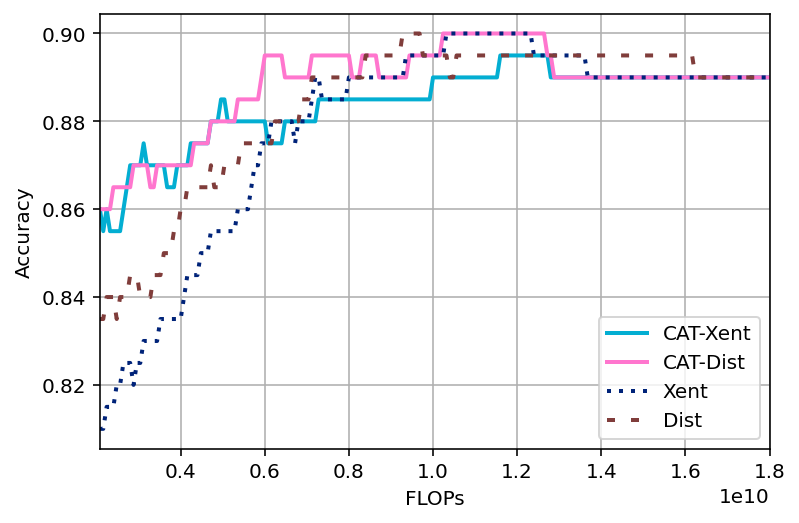}
        \subcaption{Mnli matched (200)}
    \end{minipage}
\caption{Task-specific Quality-FLOPs tradeoff on FLAN dataset with accuracy as quality measurement. Tasks 1-16 are listed here.}

\label{apdx:fig:flan_acc_task_specific_1}
\end{figure}

\begin{figure}
\renewcommand{\thesubfigure}{\arabic{subfigure}}
    \centering
    \begin{minipage}[b]{0.24\textwidth}
        \centering
        \includegraphics[width=\textwidth]{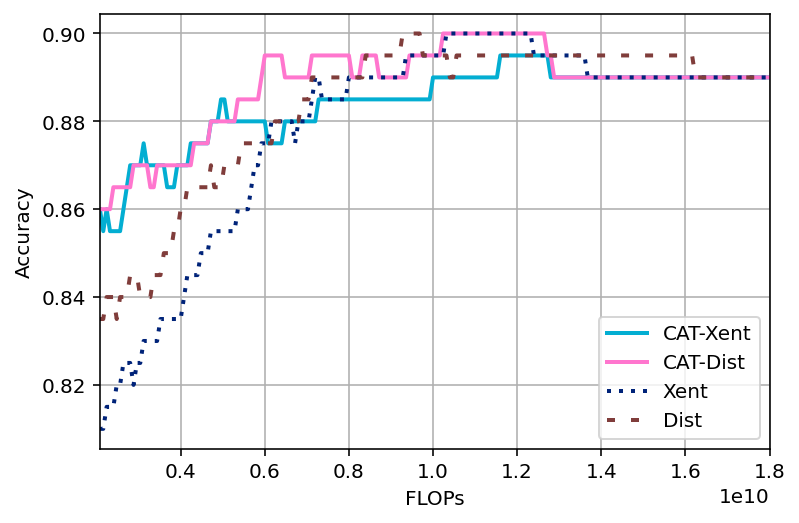}
        \subcaption{Mnli mismatched (200)}
    \end{minipage}
    \hfill
    \begin{minipage}[b]{0.24\textwidth}
        \centering
        \includegraphics[width=\textwidth]{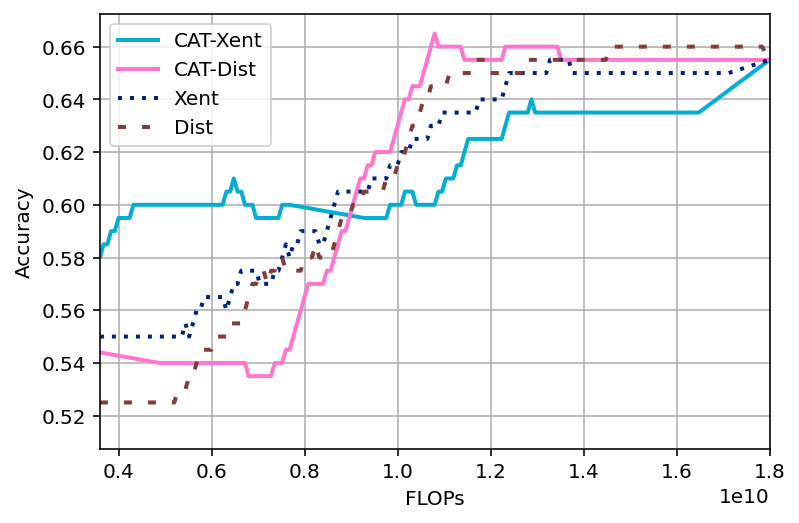}
        \subcaption{OpenBookQA (200)}
    \end{minipage}
    \hfill
    \begin{minipage}[b]{0.24\textwidth}
        \centering
        \includegraphics[width=\textwidth]{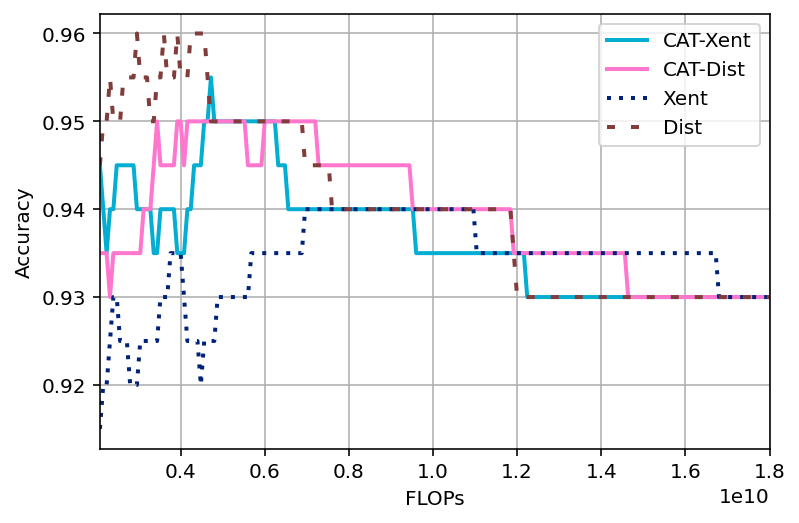}
        \subcaption{PawsWiki (200)}
    \end{minipage}
    \hfill
    \begin{minipage}[b]{0.24\textwidth}
        \centering
        \includegraphics[width=\textwidth]{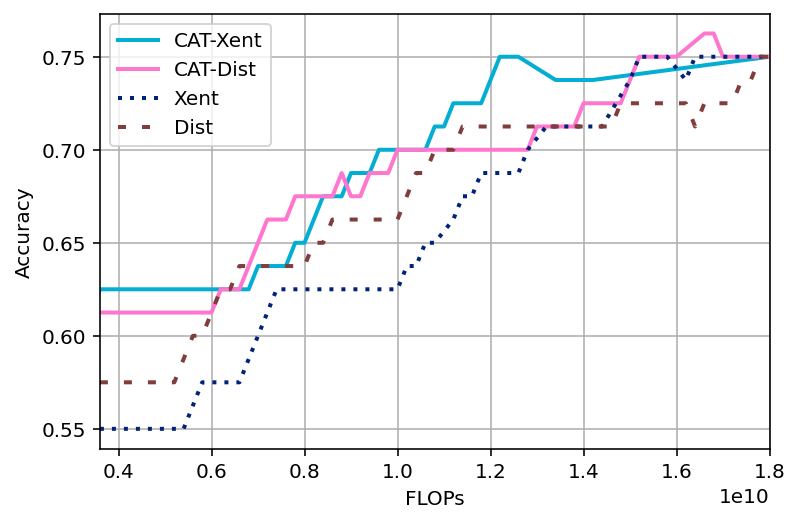}
        \subcaption{PiQA (80)}
    \end{minipage}
    \hfill
    \begin{minipage}[b]{0.24\textwidth}
        \centering
        \includegraphics[width=\textwidth]{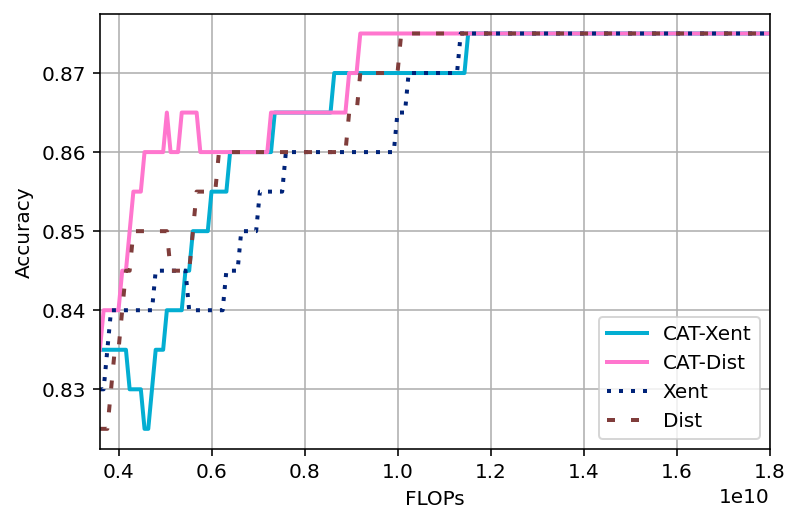}
        \subcaption{Qnli (200)}
    \end{minipage}
    \hfill
    \begin{minipage}[b]{0.24\textwidth}
        \centering
        \includegraphics[width=\textwidth]{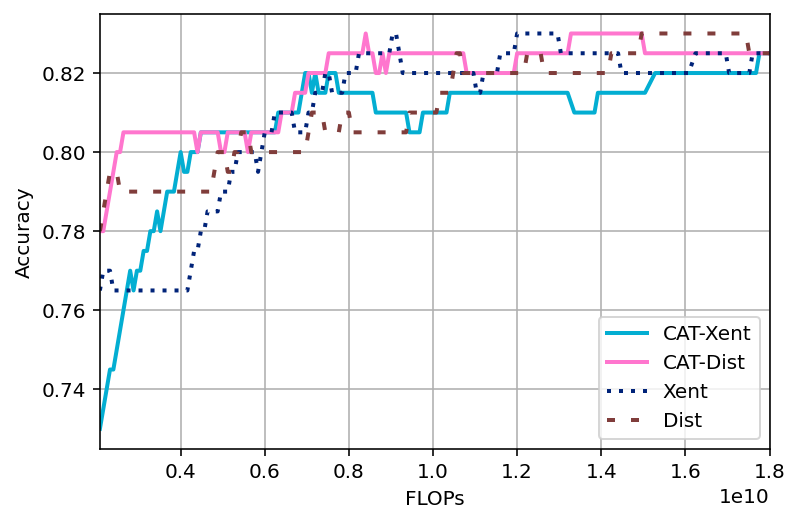}
        \subcaption{RTE (200)}
    \end{minipage}
    \hfill
    \begin{minipage}[b]{0.24\textwidth}
        \centering
        \includegraphics[width=\textwidth]{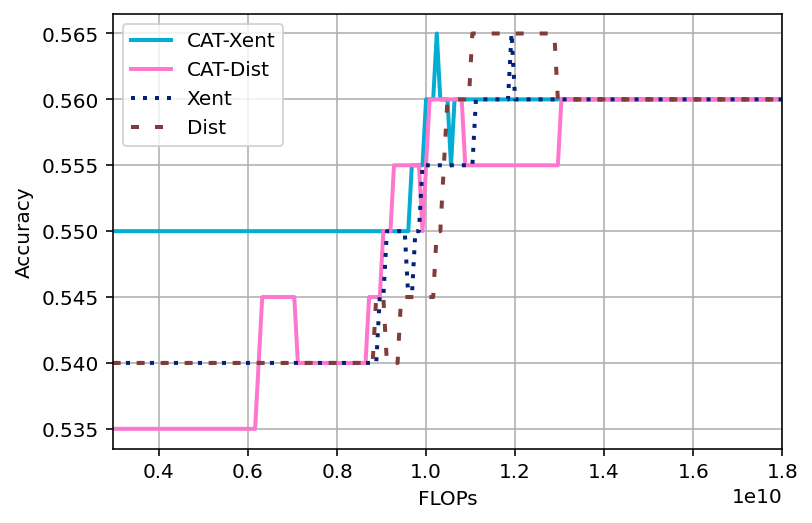}
        \subcaption{Sentiment140 (200)}
    \end{minipage}
    \hfill
    \begin{minipage}[b]{0.24\textwidth}
        \centering
        \includegraphics[width=\textwidth]{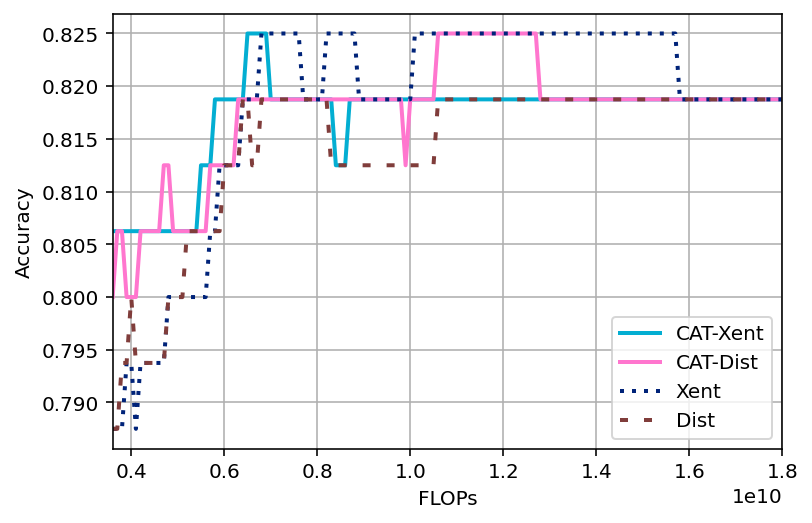}
        \subcaption{SNLI (160)}
    \end{minipage}
    \hfill
    \begin{minipage}[b]{0.24\textwidth}
        \centering
        \includegraphics[width=\textwidth]{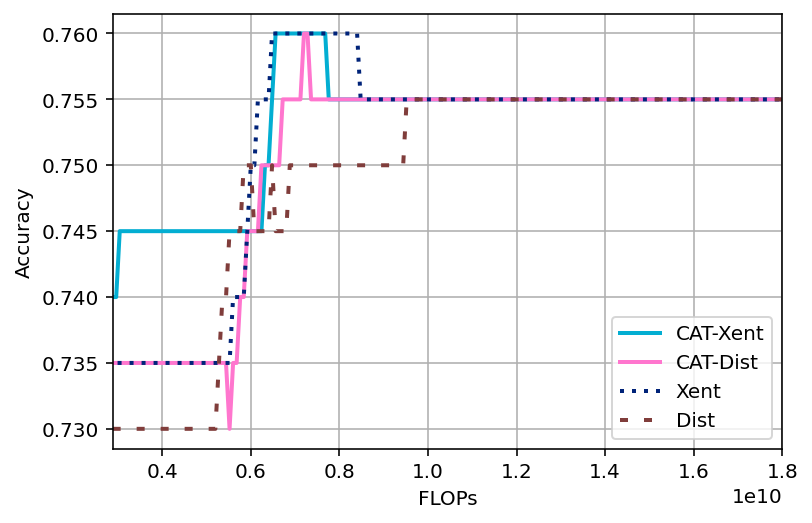}
        \subcaption{SST2 (200)}
    \end{minipage}
    \hfill
    \begin{minipage}[b]{0.24\textwidth}
        \centering
        \includegraphics[width=\textwidth]{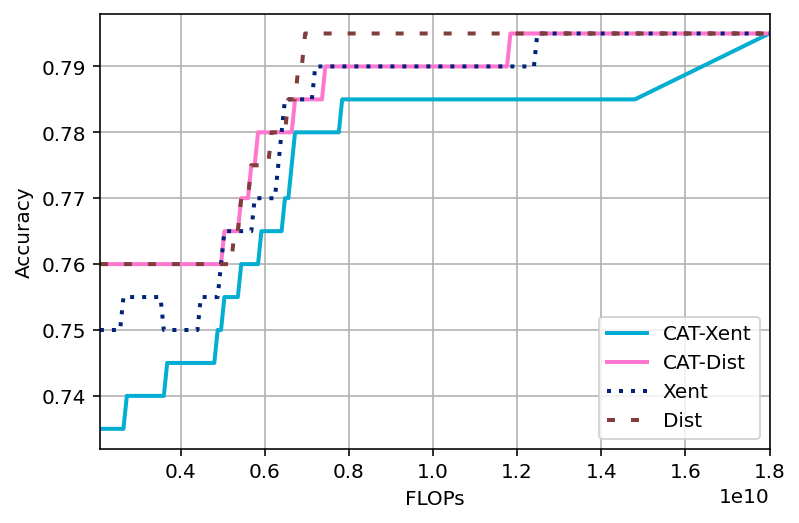}
        \subcaption{Story-cloze (200)}
    \end{minipage}
    \hfill
    \begin{minipage}[b]{0.24\textwidth}
        \centering
        \includegraphics[width=\textwidth]{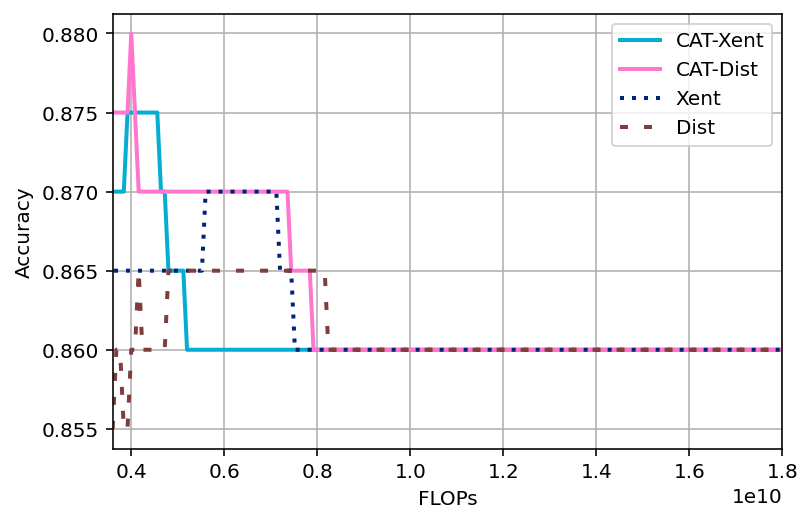}
        \subcaption{Trec (200)}
    \end{minipage}
    \hfill
    \begin{minipage}[b]{0.24\textwidth}
        \centering
        \includegraphics[width=\textwidth]{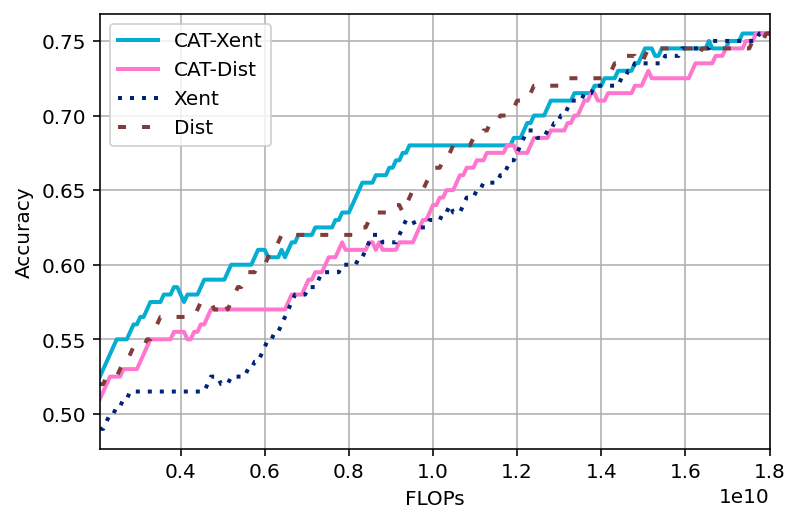}
        \subcaption{Anli R1 (200)}
    \end{minipage}
    \hfill
    \begin{minipage}[b]{0.24\textwidth}
        \centering
        \includegraphics[width=\textwidth]{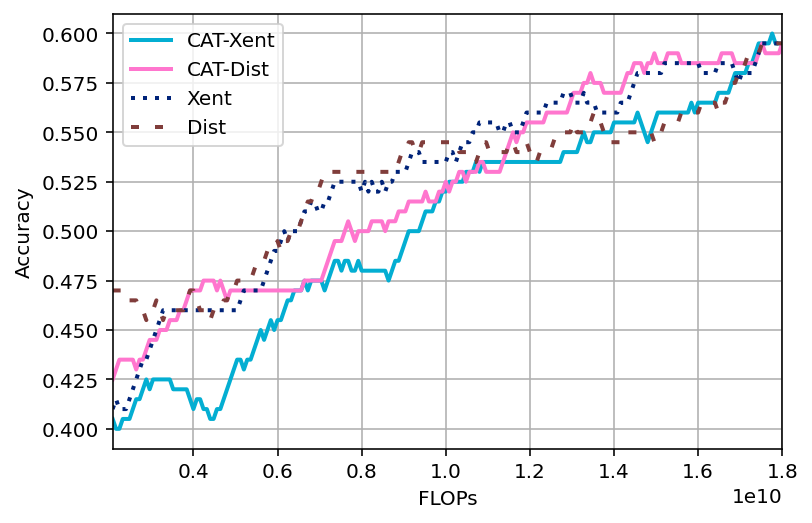}
        \subcaption{Anli R2 (200)}
    \end{minipage}
    \hfill
    \begin{minipage}[b]{0.24\textwidth}
        \centering
        \includegraphics[width=\textwidth]{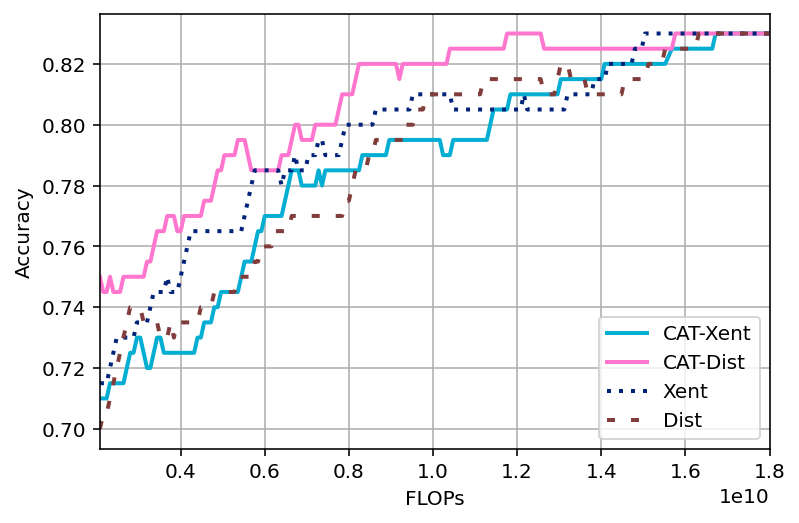}
        \subcaption{WIC (200)}
    \end{minipage}
    \hfill
    \begin{minipage}[b]{0.24\textwidth}
        \centering
        \includegraphics[width=\textwidth]{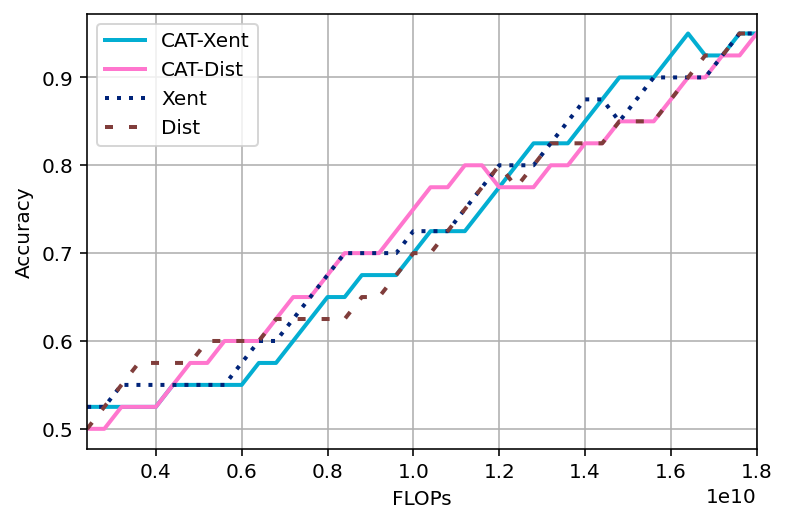}
        \subcaption{WSC (40)}
    \end{minipage}
    \hfill
    \begin{minipage}[b]{0.24\textwidth}
        \centering
        \includegraphics[width=\textwidth]{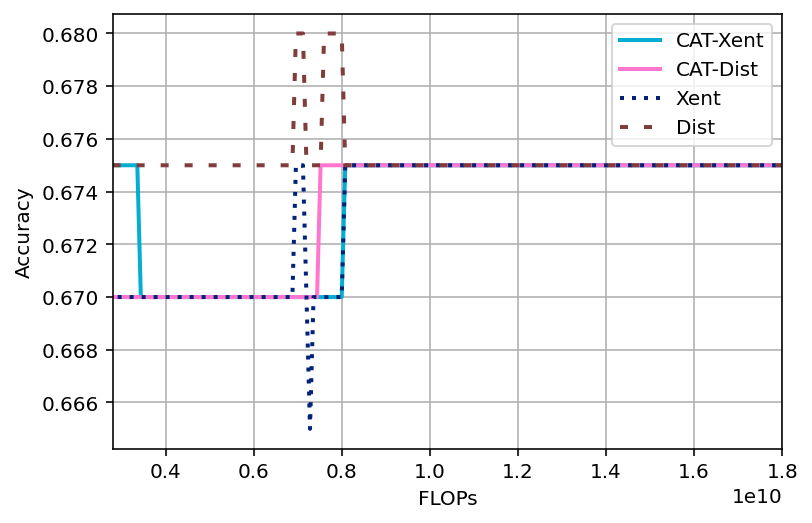}
        \subcaption{Yelp Polarity Reviews (200)}
    \end{minipage}
\caption{Task-specific Quality-FLOPs tradeoff on FLAN dataset with accuracy as quality measurement. Task 17-32 are listed here.}
\label{apdx:fig:flan_acc_task_specific_2}
\renewcommand{\thesubfigure}{\alph{subfigure}}
\end{figure}

Figures \ref{apdx:fig:sg_task_specific} and \ref{apdx:fig:wmt_task_specific} show the task-specific comparisons for each task on the SuperGLUE and WMT22 datasets. Beyond benefits to the cascade, we observed that CAT can help the small model itself perform better on some tasks, while for others, it may result in a downgrade. This aligns with our expectation that CAT filters out tokens that are difficult to predict, benefiting 'easier' tasks more. For difficult tasks, since more tokens are filtered, the model's intrinsic performance may be adversely affected. This observation is also evident in the FLAN2021 dataset, as shown in Figures \ref{apdx:fig:flan_bleu_task_specific}, \ref{apdx:fig:flan_acc_task_specific_1}, and \ref{apdx:fig:flan_acc_task_specific_2}. Note that the FLAN2021 dataset consists of 39 tasks evaluated by accuracy. We found that for 7 of these tasks (CoQA, DROP, Natural Questions, SAMSum, SQuAD v1, SQuAD v2, and TriviaQA), the four methods in the comparison showed no clear differences. We did not include plots for these 7 tasks to maintain clarity in figure arrangement.

For each task-specific plot, we included the task name and the evaluation set size in the title.

\section{CAT on model pre-training}
\label{app.cat4pretrain}

In the main body of the paper we presented CAT as a method for fine-tuning (small) language models which will target a cascaded deployment topology. One may similarly inquire whether pretraining tasks can benefit from a similar phenomenon, if it is known that the resulting models will also be deployed in a cascade.

We performed some limited exploration of CAT for pre-training of language models and did indeed see that training with CAT yielded similar benefits on next-token prediction accuracies during pretraining. We used these early experiments on pretraining as effectively an iteration ground and so were not as rigorous with our evaluations, considering the fine-tuning regime to be the more promising area--therefore we evaluated only with teacher forcing, never using full decoding.


\begin{figure}[t]
    \centering
    \includegraphics[width=0.5\textwidth]{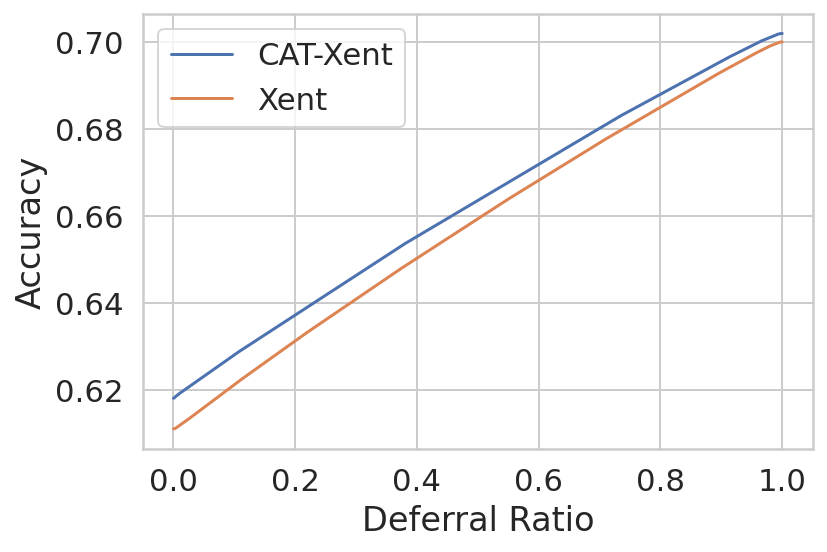}
    \label{apdx:fig:cat_pretrain_deferral}
    \caption{Next-token-prediction accuracy on C4, using \texttt{T5-base} and \texttt{T5-large} models ~\citep{raffel2020c4t5} for $p_S$ and $p_L$. Here accuracy is for the next-token prediction. Decoding performed with teacher forcing.}
    \label{apdx:fig:pretrain_deferral}
\end{figure}

The results in Figure~ ~\ref{apdx:fig:pretrain_deferral} served as essentially the starting point of the CAT workstream, demonstrating that significant gains could be had by relatively simply modification of loss functions.

\end{document}